\documentclass[11pt]{article} % For LaTeX2e

\usepackage{hyperref}
\usepackage{url}
\usepackage{multirow}

\usepackage{amsfonts,epsfig,graphicx}
\usepackage{amsmath,amssymb,amsthm}
\usepackage{fullpage}
\usepackage{hhline}
\usepackage{sidecap}

\usepackage{epsf}
\usepackage{graphics}
\usepackage{amsfonts}
\usepackage{amsmath}
\usepackage{amsthm}
\usepackage{psfrag}

\usepackage{enumitem}
\usepackage{caption}
\usepackage{subcaption}

\usepackage{xcolor,colortbl}

\usepackage{wrapfig}
\usepackage{fullpage}
\usepackage{picinpar}

\let\oldbibliography\thebibliography
\renewcommand{\thebibliography}[1]{%
  \oldbibliography{#1}%
  \setlength{\itemsep}{0pt}%
}

\newtheorem{definition}{Definition}
\newtheorem{theorem}{Theorem} 
\newtheorem{lemma}{Lemma}

\newtheorem{corollary}{Corollary}

\theoremstyle{definition}
\newtheorem{example}{Example}

\newcommand{\defn}{\ensuremath{:\,=}}

\newcommand{\Exs}{\ensuremath{\mathbb{E}}}

\newcommand{\cuplimits}{\operatornamewithlimits{\cup}}
\newcommand{\caplimits}{\operatornamewithlimits{\cap}}

\newcommand{\kl}[2]{\ensuremath{D_{\tiny{\operatorname{KL}}}(#1 \|
    #2)}}

\newcommand{\mprob}{\ensuremath{\mathbb{P}}}
\newcommand{\half}{\ensuremath{{\frac{1}{2}}}}

\newcommand{\wt}{\ensuremath{M}}

\newcommand{\numitems}{\ensuremath{n}}
\newcommand{\obs}{\ensuremath{Y}}

\newcommand{\perm}{\ensuremath{\pi}}

\newcommand{\permtilde}{\ensuremath{\widetilde{\perm}}}
\newcommand{\permstar}{\ensuremath{\perm^*}}

 \newcommand{\plaincon}{c}
\newcommand{\wtparam}{w}

\newcommand{\cardinality}[1]{\ensuremath{| #1 |}}

\newcommand{\numobs}{\numitems}

\newcommand{\score}{\tau}

\makeatletter
\long\def\@makecaption#1#2{
        \vskip 0.8ex
        \setbox\@tempboxa\hbox{\small {\bf #1:} #2}
        \parindent 1.5em  %% How can we use the global value of this???
        \dimen0=\hsize
        \advance\dimen0 by -3em
        \ifdim \wd\@tempboxa >\dimen0
                \hbox to \hsize{
                        \parindent 0em
                        \hfil 
                        \parbox{\dimen0}{\def\baselinestretch{0.96}\small
                                {\bf #1.} #2
                                } 
                        \hfil}
        \else \hbox to \hsize{\hfil \box\@tempboxa \hfil}
        \fi
        }
\makeatother

   \newcounter{parentnumber}

\newcommand{\pp}{\ensuremath{p}}

\newcommand{\matsnorm}[2]{|\!|\!| #1 | \! | \!|_{{#2}}}

\newcommand{\order}{\ensuremath{\mathcal{O}}}

\long\def\comment#1{}

\newcommand{\indicator}[1]{\ensuremath{\mathbf{1}\{#1\}}}

\newcommand{\real}{\ensuremath{\mathbb{R}}}

\newcommand{\numrepeat}{\ensuremath{r}}
\newcommand{\repvar}{\ensuremath{\ell}}

\newenvironment{carlist} {\begin{list}{$\bullet$}
    {\setlength{\topsep}{0in} \setlength{\partopsep}{0in}
      \setlength{\parsep}{0in} \setlength{\itemsep}{\parskip}
      \setlength{\leftmargin}{0.07in} \setlength{\rightmargin}{0.08in}
      \setlength{\listparindent}{0in} \setlength{\labelwidth}{0.08in}
      \setlength{\labelsep}{0.1in} \setlength{\itemindent}{0in}}}
               {\end{list}}

\newcommand{\bcar}{\begin{carlist}} \newcommand{\ecar}{\end{carlist}}

\newcommand{\KCON}{\ensuremath{c}}

\newcommand{\UUP}{\ensuremath{\KCON_u}}

\newcommand{\UNUM}{\ensuremath{\KCON_0}}

\newcommand{\mary}{L}
\newcommand{\packnum}{\ensuremath{\mary}}

\newcommand{\dham}{\ensuremath{D_{\mathrm{H}}}}
\newcommand{\hamthr}{h}
\newcommand{\hamconexp}{\nu_1}
\newcommand{\hamconmul}{\nu_2}
\newcommand{\genconexp}{\mu_1}
\newcommand{\genconmul}{\mu_2}

\newcommand{\toprank}{k}
\newcommand{\topset}{S}
\newcommand{\topstar}{\mathcal{S}^*_\toprank}

\newcommand{\Sstar}{\mathcal{S}^*}
\newcommand{\tophat}{\widehat{\mathcal{S}}_\toprank}
\newcommand{\toptilde}{\widetilde{\mathcal{S}}_\toprank}
\newcommand{\topdagger}{\mathcal{S}^\dagger_\toprank}
\newcommand{\critical}{\Delta}
\newcommand{\criticalK}{\critical_{\toprank}}
\newcommand{\criticalKham}{\critical_{\toprank, \, \hamthr}}
\newcommand{\criticalrank}{\ensuremath{\critical_0}}
\newcommand{\itemk}{(\toprank)}
\newcommand{\itemkk}{(\toprank+1)}

\newcommand{\itemup}{a}
\newcommand{\itemdown}{b}

\newcommand{\itemrv}{X}
\newcommand{\itemrvcent}{\overline{X}}

\newcommand{\compare}[2]{M_{#1 #2}}

\newcommand{\event}{\mathcal{E}}

\newcommand{\winstart}{u}
\newcommand{\winend}{v}

\newcommand{\allallowedsets}{\ensuremath{\mathfrak{S}}}
\newcommand{\allowedset}{\ensuremath{T}}
\newcommand{\alloweditem}{\ensuremath{t}}
\newcommand{\numallowedsets}{\ensuremath{\beta}}
\newcommand{\varallowedset}{b}
\newcommand{\varalloweditem}{j}
\newcommand{\epsothermetric}{\epsilon}
\newcommand{\monotone}{\Lambda}
\newcommand{\criticalS}{\critical_{\allallowedsets}}
\newcommand{\epsAlmost}{\epsilon}
\newcommand{\erased}{\phi}

\newcommand{\PLAINTEST}{\ensuremath{\phi}}
\newcommand{\TESTY}{\PLAINTEST(Y)}

\newcommand{\Pbar}{\ensuremath{\widebar{\mprob}}}

\newcommand{\KEYCON}{\ensuremath{\alpha}}
\newcommand{\FAMK}[1]{\ensuremath{\mathcal{F}_k(#1)}}
\newcommand{\FAMKH}[1]{\ensuremath{\mathcal{F}_{k,h}(#1)}}
\newcommand{\FAMSET}[1]{\ensuremath{\mathcal{F}_{\allallowedsets}(#1)}}
\newcommand{\FAMGEN}[2]{\ensuremath{\mathcal{F}_{#2}(#1)}}

\newlength{\widebarargwidth}
\newlength{\widebarargheight}
\newlength{\widebarargdepth}
\DeclareRobustCommand{\widebar}[1]{%
  \settowidth{\widebarargwidth}{\ensuremath{#1}}%
  \settoheight{\widebarargheight}{\ensuremath{#1}}%
  \settodepth{\widebarargdepth}{\ensuremath{#1}}%
  \addtolength{\widebarargwidth}{-0.3\widebarargheight}%
  \addtolength{\widebarargwidth}{-0.3\widebarargdepth}%
  \makebox[0pt][l]{\hspace{0.3\widebarargheight}%
    \hspace{0.3\widebarargdepth}%
    \addtolength{\widebarargheight}{0.3ex}%
    \rule[\widebarargheight]{0.95\widebarargwidth}{0.1ex}}%
  {#1}}

\newcommand{\vtiny}{\vspace*{.1in}}

\newcommand{\Mmat}{\ensuremath{M}}
\newcommand{\ModelProb}{\ensuremath{\mprob_M}}

\newcommand{\OrderCompare}[2]{\ensuremath{\Mmat_{(#1) #2}}}

\begin{document}

\begin{center} {\LARGE{\bf{
Simple, Robust and Optimal Ranking\\ from Pairwise Comparisons} }}

\vspace*{.3in}

\begin{tabular}{ccc}
Nihar B. Shah$^\dagger$ & & Martin J. Wainwright$^{\dagger, \ast}$
\\ {\tt nihar@eecs.berkeley.edu} & & {\tt wainwrig@berkeley.edu}
\end{tabular}

\vspace*{.2in}
{\large{
\begin{tabular}{c}
Department of EECS$^\dagger$, and Department of Statistics$^\ast$
\\ University of California, Berkeley \\ Berkeley, CA 94720
\end{tabular}
\vspace*{.2in} }}

%\today

\vspace*{.2in}

\begin{abstract}
We consider data in the form of pairwise comparisons of $\numitems$
items, with the goal of precisely identifying the top $\toprank$ items
for some value of $\toprank < \numitems$, or alternatively, recovering
a ranking of all the items.  We analyze the Copeland counting
algorithm that ranks the items in order of the number of pairwise
comparisons won, and show it has three attractive features: (a) its
computational efficiency leads to speed-ups of several orders of
magnitude in computation time as compared to prior work; (b) it is
robust in that theoretical guarantees impose no conditions on the
underlying matrix of pairwise-comparison probabilities, in contrast to
some prior work that applies only to the BTL parametric model; and (c)
it is an optimal method up to constant factors, meaning that it
achieves the information-theoretic limits for recovering the top
$\toprank$-subset. We extend our results to obtain sharp guarantees
for approximate recovery under the Hamming distortion metric, and more
generally, to any arbitrary error requirement that satisfies a simple
and natural monotonicity condition.
\end{abstract}
\end{center}

%%%%%%%%%%%%%%%%%%%%%%%%%%%%%%%%%%%%%%%%%%%%%%%%%%%%%%%%%%%%%%%%%%%%%%%%%

\section{Introduction}
\label{SecIntroduction}
Ranking problems involve a collection of $\numitems$ items, and some
unknown underlying total ordering of these items.  In many
applications, one may observe (noisy) comparisons between various
pairs of items. Examples include matches between football teams in
tournament play; consumer's preference ratings in marketing; and
certain types of voting systems in politics.  Given a set of such
noisy comparisons between items, it is often of interest to find the
true underlying ordering of all $\numitems$ items, or alternatively,
given some given positive integer $\toprank < \numitems$, to find the
subset of $\toprank$ most highly rated items.  These two problems are
the focus of this paper.

There is a substantial literature on the problem of finding
approximate rankings based on noisy pairwise comparisons.  A number of
papers (e.g.,~\cite{kenyon2007rank, braverman2008noisy,
  eriksson2013learning}) consider models in which the probability of a
pairwise comparison agreeing with the underlying order is identical
across all pairs. These results break down when for one or more pairs,
the probability of agreeing with the underlying ranking is either comes close to or is exactly equal to $\half$.  Another set of
papers~\cite{hunter2004mm, negahban2012iterative, hajek2014minimax,
  soufiani2014computing, shah2015estimation} work using parametric
models of pairwise comparisons, and address the problem of recovering
the parameters associated to every individual item. A more recent
line of work~\cite{chatterjee2014matrix, shah2015stochastically,
  shah2016feeling} studies a more general class of models based on the
notion of strong stochastic transitivity (SST), and derives conditions
on recovering the pairwise comparison probabilities
themselves. However, it remains unclear whether or not these results
can directly extend to tight bounds for the problem of recovery of the
top $\toprank$ items. The
works~\cite{jagabathula2008inferring,mitliagkas2011user,ammar2012efficient,
  ding2014topic} consider mixture models, in which every pairwise
comparison is associated to a certain individual making the
comparison, and it is assumed that the preferences across individuals
can be described by a low-dimensional model.

Most related to our work are the papers~\cite{wauthier2013efficient,
  rajkumar2014statistical, rajkumar2015ranking, chen2015spectral},
which we discuss in more detail here. Wauthier et
al.~\cite{wauthier2013efficient} analyze a weighted counting algorithm
to recover approximate rankings; their analysis applies to a specific
model in which the pairwise comparison between any pair of items
remains faithful to their relative positions in the true ranking with
a probability common across all pairs.  They consider recovery of an
approximate ranking (under Kendall's tau and maximum displacement
metrics), but do not provide results on exact recovery.  As the
analysis of this paper shows, their bounds are quite loose: their
results are tight only when there are a total of at least
$\Theta(\numitems^2)$ comparisons. The pair of
papers~\cite{rajkumar2014statistical, rajkumar2015ranking} by Rajkumar
et al. consider ranking under several models and several metrics. In
the part that is common with our setting, they show that the counting
algorithm is consistent in terms of recovering the full ranking, which
automatically implies consistency in exactly recovering the top
$\toprank$ items. They obtain upper bounds on the sample complexity in
terms of a separation threshold that is identical to a parameter
$\criticalK$ defined subsequently in this paper (see
Section~\ref{SecResults}).  However, as our analysis shows, their
bounds are loose by at least an order of magnitude. They also assume a
certain high-SNR condition on the probabilities,  an assumption that
is not imposed in our analysis.

Finally, in very recent work on this problem, Chen and
Suh~\cite{chen2015spectral} proposed an algorithm called the Spectral
MLE for exact recovery of the top $\toprank$ items.  They showed that,
if the pairwise observations are assumed to drawn according to the
Bradley-Terry-Luce (BTL) parametric model~\cite{bradley1952rank,
  luce1959individual}, the Spectral MLE algorithm recovers the
$\toprank$ items correctly with high probability under certain
regularity conditions.  In addition, they also show, via matching
lower bounds, that their regularity conditions are tight up to
constant factors.  While these guarantees are attractive, it is
natural to ask how such an algorithm behaves when the data is not
drawn from the BTL model.  In real-world instances of pairwise ranking
data, it is often found that parametric models, such as the BTL model
and its variants, fail to provide accurate fits (for instance, see the
papers~\cite{davidson1959experimental, mclaughlin1965stochastic,
  tversky1972elimination, ballinger1997decisions} and references
therein).

With this context, the main contribution of this paper is to analyze a
classical counting-based method for ranking, often called the Copeland
method~\cite{copeland1951reasonable}, and to show that it is simple,
optimal and robust.  Our analysis does not require that the
data-generating mechanism follow either the BTL or other parametric
assumptions, nor other regularity conditions such as stochastic
transitivity. We show that the Copeland counting algorithm has the
following properties:
\begin{itemize}[leftmargin=*]
\item Simplicity: The algorithm is simple, as it just orders the items
  by the number of pairwise comparisons won. As we will subsequently
  see, the execution time of this counting algorithm is several orders
  of magnitude lower as compared to prior work.
\item Optimality: We derive conditions under which the counting
  algorithm achieves the stated goals, and by means of matching
  information-theoretic lower bounds, show that these conditions are
  tight.
\item Robustness: The guarantees that we prove do not require any
  assumptions on the pairwise-comparison probabilities, and the counting 
  algorithm performs well for various classes of data sets.  In
  contrast, we find that the spectral MLE algorithm performs poorly
  when the data is not drawn from the BTL model.
\end{itemize}

In doing so, we consider three different instantiations of the problem
of set-based recovery: (i) Recovering the top $\toprank$ items
perfectly; (ii) Recovering the top $\toprank$ items allowing for a
certain Hamming error tolerance; and (iii) a more general recovery
problem for set families that satisfy a natural ``set-monotonicity''
condition.  In order to tackle this third problem, we introduce a
general framework that allows us to treat a variety of problems in the
literature in an unified manner.

The remainder of this paper is organized as follows.  We begin in
Section~\ref{SecBackground} with background and a more precise
formulation of the problem. Section~\ref{SecResults} presents our main
theoretical results on top-$\toprank$ recovery under various
requirements. Section~\ref{SecSimulations} provides the results of
experiments on both simulated and real-world data sets.  We provide
all proofs in Section~\ref{SecProofs}. The paper concludes with a
discussion in Section~\ref{SecDiscussion}.

%%%%%%%%%%%%%%%%%%%%%%%%%%%%%%%%%%%%%%%%%%%%%%%%%%%%%%%%%%%%%%%%%%%%%%%%

\section{Background and problem formulation}
\label{SecBackground}

In this section, we provide a more formal statement of the problem
along with background on various types of ranking models.

%%%%%%%%%%%%%%%%%%%%%%%%%%%%%%%%%%%%%%%%%%%%%%%

\subsection{Problem statement}

Given an integer $\numitems \geq 2$, we consider a collection of
$\numitems$ items, indexed by the set \mbox{$[\numitems] \defn \{1,
  \ldots, \numitems \}$.}  For each pair $i \neq j$, we let
$\compare{i}{j}$ denote the probability that item $i$ wins the
comparison with item $j$.  We assume that 
that each comparison necessarily results
in one winner, meaning that
\begin{align}
\label{EqnBase}
\compare{i}{j} + \compare{j}{i} = 1, \qquad \mbox{and} \quad
\compare{i}{i} = \half,
\end{align}
where we set the diagonal for concreteness.

For any item $i \in [\numitems]$, we define an associated score
$\score_i$ as
\begin{align} 
\label{eq:defn_scores}
\score_i & \defn \frac{1}{\numitems} \sum_{j = 1}^{\numitems}
\compare{i}{j}.
\end{align}
In words, the score $\score_i$ of any item $i \in [\numitems]$
corresponds to the probability that item $i$ beats an item chosen
uniformly at random from all $\numitems$ items.

Given a set of noisy pairwise comparisons, our goals are (a) to
recover the $\toprank$ items with the maximum values of their scores;
and (b) to recover the full ordering of all the items as defined by
the score vector.  The notion of ranking items via their
scores~\eqref{eq:defn_scores} generalizes the explicit rankings under
popular models in the literature. Indeed, as we discuss shortly, most
models of pairwise comparisons considered in the literature either
implicitly or explicitly assume that the items are ranked according to
their scores.  Note that neither the scores $\{\score_i\}_{i \in
  [\numitems]}$ nor the matrix $\Mmat \defn \{\compare{i}{j}\}_{i,j
  \in [\numitems]}$ of probabilities are assumed to be known.

More concretely, we consider a random-design observation model defined
as follows. Each pair is associated with a random number of noisy
comparisons, following a binomial distribution with parameters
$(\numrepeat, \pp)$, where $\numrepeat \geq 1$ is the number of trials
and $\pp \in (0,1]$ is the probability of making a comparison on any
  given trial.  Thus, each pair $(i,j)$ is associated with a binomial
  random variable with parameters $(\numrepeat, \pp)$ that governs the
  number of comparisons between the pair of items.  We assume that the
  observation sequences for different pairs are independent.  Note
  that in the special case $\pp = 1$, this random binomial model
  reduces to the case in which we observe exactly $\numrepeat$
  observations of each pair; in the special case $\numrepeat=1$, the set of pairs compared form an $(\numitems, \pp)$ Erd\H{o}s-R\'enyi random graph.

In this paper, we begin in Section~\ref{SecExact} by analyzing the
problem of exact recovery.  More precisely, for a given matrix $\Mmat$
of pairwise probabilities, suppose that we let $\topstar$ denote the
(unknown) set of $\toprank$ items with the largest values of their
respective scores, assumed to be unique for concreteness.

Given noisy observations specified by the pairwise probabilities
$\Mmat$, our goal is to establish conditions under which there exists
some algorithm $\tophat$ that identifies $\toprank$ items based on the
outcomes of various comparisons such that the probability $\ModelProb
( \tophat = \topstar)$ is very close to one. In the case of recovering
the full ranking, our goal is to identify conditions that ensure that
the probability $\ModelProb \big(\caplimits \limits_{\toprank \in [\numitems]} (\tophat =
  \topstar)\big)$ is close to one.

In Section~\ref{SecHamming}, we consider the problem of recovering a
set of $\toprank$ items that approximates $\topstar$ with a minimal
Hamming error For any two subsets of $[\numitems]$, we define their
Hamming distance $\dham$, also referred to as their Hamming error, to
be the number of items that belong to exactly one of the two
sets---that is
\begin{align}
\label{EqnDefnHammingSet}
\dham(A, B) & = \mbox{card} \Big( \{ A \cup B \} \backslash \{A \cap B
\} \Big).
\end{align}
For a given user-defined tolerance parameter $\hamthr \geq 0$, we
derive conditions that ensure that \mbox{$\dham(\tophat, \topstar)
  \leq 2\hamthr$} with high probability.

Finally, we generalize our results to the problem of satisfying any a
general class of requirements on set families.  These requirement are
specified in terms of which $\toprank$-sized subsets of the items are
allowed, and is required to satisfy only one natural condition, that
of set-monotonicity, meaning that replacing an item in an allowed set
with a higher rank item should also be allowed.  See
Section~\ref{SecGeneralReq} for more details on this general
framework.

%%%%%%%%%%%%%%%%%%%%%%%%%%%%%%%%%%%%%%%%%%%%%%%%%%%%%%%%%%%%%%%%%%%%%%%

\subsection{A range of pairwise comparison models} 

To be clear, our work makes no assumptions on the form of the pairwise
comparison probabilities.  However, so as to put our work in context
of the literature, let us briefly review some standard models uesd for
pairwise comparison data.

\paragraph{Parametric models:} 
A broad class of parametric models, including the Bradley-Terry-Luce
(BTL) model as a special case~\cite{bradley1952rank,
  luce1959individual}, are based on assuming the existence of
``quality'' parameter $\wtparam_i \in \real$ for each item $i$, and
requiring that the probability of an item beating another is a
specific function of the difference between their values.  In the BTL
model, the probability $\compare{i}{j}$ that $i$ beats $j$ is given by
the logistic model
\begin{subequations}
\begin{align}
\compare{i}{j} = \frac{1}{1 + e^{-(\wtparam_i - \wtparam_j)}}.
\label{eq:defn_BTL}
\end{align}
More generally, parametric models assume that the pairwise comparison
probabilities take the form
\begin{align}
\compare{i}{j} =
F(\wtparam_i - \wtparam_j), \label{eq:defn_parametric} 
\end{align} 
\end{subequations}
where $F: \real \rightarrow [0,1]$ is some strictly increasing
cumulative distribution function.

By construction, any parametric model has the following property: if
$\wtparam_i > \wtparam_j$ for some pair of items $(i,j)$, then we are
also guaranteed that $\compare{i}{\ell} > \compare{j}{\ell}$ for every
item $\ell$.  As a consequence, we are guaranteed that $\score_i >
\score_j$, which implies that ordering of the items in terms of their
quality vector $\wtparam \in \real^\numobs$ is identical to their
ordering in terms of the score vector $\score \in \real^\numitems$.
Consequently, if the data is actually drawn from a parametric model,
then recovering the top $\toprank$ items according to their scores is
the same as recovering the top $\toprank$ items according their
respective quality parameters.

\paragraph{Strong Stochastic Transitivity (SST) class:}  The class of 
strong stochastic transitivity (SST) models is a superset of
parametric models~\cite{shah2015stochastically}. It does not assume
the existence of a quality vector, nor does it assume any specific
form of the probabilities as in equation~\eqref{eq:defn_BTL}. Instead,
the SST class is defined by assuming the existence of a total ordering
of the $\numitems$ items, and imposing the inequality constraints
$\compare{i}{\ell} \geq \compare{j}{\ell}$ for every pair of items
$(i,j)$ where $i$ is ranked above $j$ in the ordering, and every item
$\ell$. One can verify that an ordering by the scores $\{ \score_i
\}_{i \in [\numitems]}$ of the items lead to an ordering of the items
that is consistent with that defined by the SST class. \\

Thus, we see that in a broad class of models for pairwise ranking, the
total ordering defined by the score vectors~\eqref{eq:defn_scores}
coincides with the underlying ordering used to define the models.  In
this paper, we analyze the performance of a counting algorithm,
without imposing any modeling conditions on the family of pairwise
probabilities. The next three sections establish theoretical
guarantees on the recovery of the top $\toprank$ items under various
requirements.

%%%%%%%%%%%%%%%%%%%%%%%%%%%%%%%%%%%%%%%%%%%%%%%%%%%%%%%%%%%%%%%%%%%%%%%%%%%%

\subsection{Copeland counting algorithm}

The analysis of this paper focuses on a simple counting-based
algorithm, often called the Copeland
method~\cite{copeland1951reasonable}.  It can be also be viewed as a
special case of the Borda count method~\cite{de1781memoire}, which
applies more generally to observations that consist of rankings of two or
more items.  Here we describe how this method applies to the
random-design observation model introduced earlier.

More precisely, for each distinct $i,j \in [\numitems]$ and every
integer $\ell \in [\numrepeat]$, let $\obs_{ij}^{\ell} \in \{-1,0,
+1\}$ represent the outcome of the $\ell^{th}$ comparison between the
pair $i$ and $j$, defined as
\begin{align}
\obs_{ij}^\ell & = \begin{cases} 0 & \mbox{no comparison between
    $(i.j)$ in trial $\ell$} \\
+ 1 & \mbox{if comparison is made and item $i$ beats $j$} \\
-1 & \mbox{if comparison is made and item $j$ beats $i$.}
\end{cases}
\end{align}
Note that this definition ensures that $\obs_{ij}^{\ell} = -
\obs_{ji}^{\ell}$.  For $i \in [\numitems]$, the quantity
\begin{align}
N_i & \defn \sum_{j \in [\numitems]} \sum_{\ell \in [\numrepeat]}
\indicator{\obs_{ij}^{\ell} = 1}
\end{align}
corresponds to the number of pairwise comparisons won by item
$i$. Here we use $\indicator{\cdot}$ to denote the indicator function
that takes the value $1$ if its argument is true, and the value $0$
otherwise.  For each integer $k$, the vector $\{N_i\}_{i=1}^\numobs$
of number of pairwise wins defines a $k$-sized subset
\begin{align}
\label{EqnMaxPairwise}
\toptilde & = \Big \{ i \in [\numitems] \, \mid \, \mbox{$N_i$ is
  among the $k$ highest number of pairwise wins} \Big \},
\end{align}
corresponding to the set of $\toprank$ items with the largest values
of $N_i$.  Otherwise stated, the set $\toptilde$ corresponds to the
rank statistics of the top $k$-items in the pairwise win ordering. (If
there are any ties, we resolve them by choosing the indices with the
smallest value of $i$.)

%%%%%%%%%%%%%%%%%%%%%%%%%%%%%%%%%%%%%%%%%%%%%%%%%%%%%%%%%%%%%%%%%%%%%%%%%%%

\section{Main results}
\label{SecResults}

In this section, we present our main theoretical results on
top-$\toprank$ recovery under the three settings described
earlier. Note that the three settings are ordered in terms
of increasing generality, with the advantage that the
least general setting leads to the simplest form of
theoretical claim.

%%%%%%%%%%%%%%%%%%%%%%%%%%%%%%%%%%%%%%%%%%%%%%%%%%%%%%%%%%%%%%%%%%%%%%%%%

\subsection{Thresholds for exact recovery of the top $\toprank$ items}
\label{SecExact}

We begin with the goal of exactly recovering the $\toprank$ top-ranked
items.  As one might expect, the difficulty of this problem turns out
to depend on the degree of separation between the top $\toprank$ items
and the remaining items.  More precisely, let us use $\itemk$ and
$\itemkk$ to denote the indices of the items that are ranked
$\toprank^{th}$ and $(\toprank +1)^{th}$ respectively.  With this
notation, the \emph{$\toprank$-separation threshold} $\criticalK$ is
given by

\begin{align}
\label{eq:defn_criticalK}
\criticalK \defn \score_{\itemk} - \score_{\itemkk} =
\frac{1}{\numitems} \sum_{i = 1}^{\numitems} \OrderCompare{k}{i} -
\frac{1}{\numitems} \sum_{i = 1}^{\numitems} \OrderCompare{k+1}{i}.
\end{align}
In words, the quantity $\criticalK$ is the difference in the
probability of item $\itemk$ beating another item chosen uniformly at
random, versus the same probability for item $\itemkk$.

As shown by the following theorem, success or failure in recovering
the top $\toprank$ entries is determined by the size of $\criticalK$
relative to the number of items $\numitems$, observation probability
$\pp$ and number of repetitions $\numrepeat$.  In particular, consider
the family of matrices
\begin{align}
\label{EqnDefnFamk}
\FAMK{\KEYCON; \numitems, \pp, \numrepeat} & \defn \left\{ \Mmat \in
     [0,1]^{\numitems \times \numitems} \mid \, \Mmat + \Mmat^T = 1
     1^T, \mbox{ and } \criticalK \geq \KEYCON \sqrt{\frac{\log
         \numitems}{\numitems \pp  \numrepeat }} \right\}.
\end{align}
To simplify notation, we often adopt $\FAMK{\KEYCON}$ as a convenient
shorthand for this set, where its dependence on $(\numitems, \pp,
\numrepeat)$ should be understood implicitly.

With this notation, the achievable result in part (a) of the following
theorem is based on the estimator that returns the set $\toptilde$ of
the the $\toprank$ items defined by the number of pairwise comparisons
won, as defined in equation~\eqref{EqnMaxPairwise}.  On the other
hand, the lower bound in part (b) applies to \emph{any estimator},
meaning any measurable function of the observations.
\begin{theorem}
\label{thm:topk}
\begin{enumerate}
\item[(a)] For any $\KEYCON \geq 8$, the maximum pairwise win
  estimator $\toptilde$ from equation~\eqref{EqnMaxPairwise} satisfies
\begin{subequations}
\begin{align}
\label{eq:topk_tail}
\sup_{\Mmat \in \FAMK{\KEYCON}} \ModelProb \big[ \toptilde \neq
  \topstar \big] & \leq \frac{1}{\numitems^{14}}.
\end{align}
\item[(b)] Conversely, suppose that $\numitems \geq 7$ and $\pp \geq
  \frac{\log \numitems}{2 \numitems \numrepeat}$.  Then for any
  $\KEYCON \leq \frac{1}{7}$, the error probability of \emph{any}
  estimator $\tophat$ is lower bounded as
\begin{align}
\sup_{\Mmat \in \FAMK{\KEYCON}} \ModelProb \big[ \tophat \neq \topstar
  \big] & \geq \frac{1}{7}.
\end{align}
\end{subequations}
\end{enumerate}
\end{theorem}

\noindent {\bf{Remarks:}} First, it is important to note that the
negative result in part (b) holds even if the supremum is further
restricted to a particular parametric sub-class of $\FAMK{\KEYCON}$,
such as the pairwise comparison matrices generated by the BTL model, or by the SST model.
Our proof of the lower bound for exact recovery is based on a
generalization of a construction introduced by Chen and
Suh~\cite{chen2015spectral}, one adapted to the general
definition~\eqref{eq:defn_criticalK} of the separation threshold
$\criticalK$.

Second, we note that in the regime $\pp < \frac{\log \numitems}{2
  \numitems \numrepeat}$, standard results from random graph
theory~\cite{erdos1960evolution} can be used to show that there are at
least $\sqrt{\numitems}$ items (in expectation) that are never
compared to any other item.  Of course, estimating the rank is
impossible in this pathological case, so we omit it from
consideration.

Third, the two parts of the theorem in conjunction show that the
counting algorithm is essentially optimal.  The only room for
improvement is in the difference between the value $8$ of $\KEYCON$ in the
achievable result, and the value $\frac{1}{7}$ in the lower bound.\\

\vspace*{.05in}

Theorem~\ref{thm:topk} can also be used to derive guarantees for
recovery of other functions of the underlying ranking. Here we
consider the problem of identifying the ranking of all $\numitems$
items, say denoted by the permutation $\permstar$. In this case, we
require that each of the separations
$\{\critical_j\}_{j=1}^{\numobs-1}$ are suitably lower bounded: more
precisely, we study models $\Mmat$ that belong to the intersection
$\cap_{j=1}^{\numobs-1} \FAMGEN{\gamma}{j}$.

\begin{corollary}
\label{cor:rank}
Let $\permtilde$ be the permutation of the items specified by the
number of pairwise comparisons won.  Then for any $\KEYCON \geq 8$, we
have
\begin{align*}
\sup_{\Mmat \in \cap_{j=1}^{\numobs-1} \FAMGEN{\KEYCON}{j}} \ModelProb
\big[ \permtilde \neq \permstar \big] \leq \frac{1}{\numitems^{13}}.
\end{align*}
Moreover, the separation condition on
$\{\critical_j\}_{j=1}^{\numitems-1}$ that defines the set
$\cap_{j=1}^{\numitems-1} \FAMGEN{\KEYCON}{j}$ is unimprovable beyond
constant factors.
\end{corollary}
\noindent This corollary follows from the equivalence between correct
recovery of the ranking and recovering the top $\toprank$ items for
every value of $\toprank \in [\numitems]$.

%%%%%%%%%%%%%%%%%%%%%%%%%%%%%%%%%%%%%%%%%%%%%%%%%%%%%%%%%%%%%%%%%%%%%%%%%

\paragraph*{Detailed comparison to related work:}

In the remainder of this subsection, we make a detailed comparison to
the related works~\cite{wauthier2013efficient,
  rajkumar2014statistical, rajkumar2015ranking, chen2015spectral} that
we briefly discussed earlier in Section~\ref{SecIntroduction}.

Wauthier et al.~\cite{wauthier2013efficient} analyze a weighted
counting algorithm for approximate recovery of rankings; they work
under a model in which $\Mmat_{ij} = \half + \gamma$ whenever item $i$
is ranked above item $j$ in an assumed underlying ordering. Here the
parameter $\gamma \in (0, \half]$ is independent of $(i,j)$, and as a
  consequence, the best ranked item is assumed to be as likely to meet
  the worst item as it is to beat the second ranked item, for
  instance. They analyze approximate ranking under Kendall tau and
  maximum displacement metrics.  In order to have a displacement upper
  bounded by by some $\delta > 0$, their bounds require the order of
  $\frac{\numitems^5}{\delta^2 \gamma^2}$ pairwise comparisons. In
  comparison, our model is more general in that we do not impose the
  $\gamma$-condition on the pairwise probabiltiies.  When specialized
  to the $\gamma$-model, the quantities $\{
  \critical_j\}_{j=1}^\numitems$ in our analysis takes the form
  $\critical_j = \frac{2\gamma}{\numitems}$, and
  Corollary~\ref{cor:rank} shows that $\frac{\numitems \log
    \numitems}{\min_{j \in [\numitems]} \critical_j^2} =
  \frac{\numitems^3 \log \numitems}{\gamma^2}$ observations are
  sufficient to recover the exact total ordering.  Thus, for any
  constant $\delta$, Corollary~\ref{cor:rank} guarantees recover with
  a multiplicative factor of order $\frac{\numitems^2}{\log
    \numitems}$ smaller than that established by Wauthier et
  al.~\cite{wauthier2013efficient}.

The pair of papers~\cite{rajkumar2014statistical, rajkumar2015ranking}
by Rajkumar et al. consider ranking under several models and several
metrics. For the subset of their models common with our
setting---namely, Bradley-Terry-Luce and the so-called low noise
models---they show that the counting algorithm is consistent in terms
of recovering the full ranking or the top subset of items. The
guarantees are obtained under a low-noise assumpotion: namely, that
the probability of any item $i$ beating $j$ is at least $\half +
\gamma$ whenever item $i$ is ranked higher than item $j$ in an assumed
underlying ordering. Their guarantees are based on a sample size of at
least $\frac{\log \numitems}{\gamma^2 \mu^2}$, where $\mu$ is a
parameter lower bounded as $\mu \geq \frac{1}{\numitems^2}$. Once
again, our setting allows for the parameter $\gamma$ to be arbitrarily
close to zero, and furthermore as one can see from the discussion
above, our bounds are much stronger. Moreover, while Rajkumar et
al. focus on upper bounds alone, we also prove matching lower bounds
on sample complexity showing that our results are unimprovable beyond
constant factors.  It should be noted that Rajkumar et al. also
provide results for other types of ranking problems that lie outside
the class of models treated in the current paper.

Most recently, Chen and Suh~\cite{chen2015spectral} show that if the
pairwise observations are assumed to drawn according to the
Bradley-Terry-Luce (BTL) parametric model~\eqref{eq:defn_BTL}, then
their proposed Spectral MLE algorithm recovers the $\toprank$ items
correctly with high probability when a certain separation condition on
the parameters $\{\wtparam_i\}_{i=1}^{\numitems}$ of the BTL model is
satisfied.  In addition, they also show, via matching lower bounds,
that this separation condition are tight up to constant factors. In
real-world instances of pairwise ranking data, it is often found that
parametric models, such as the BTL model and its variants, fail to
provide accurate fits~\cite{davidson1959experimental,
  mclaughlin1965stochastic, tversky1972elimination,
  ballinger1997decisions}. Our results make no such assumptions on the
noise, and furthermore, our notion of the ordering of the items in
terms of their scores~\eqref{eq:defn_scores} strictly generalizes the
notion of the ordering with respect to the BTL parameters. In
empirical evaluations presented subsequently, we see that the counting
algorithm is significantly more robust to various kinds of noise, and
takes several orders of magnitude lesser time to compute.

Finally, in addition to the notion of exact recovery considered so
far, in the next two subsections we also derive tight guarantees for
the Hamming error metric and more general metrics inspired by the
requirements of many relevant
applications~\cite{ilyas2008survey,michel2005klee,babcock2003distributed,
  metwally2005efficient, kimelfeld2006finding, fagin2003optimal}.

%%%%%%%%%%%%%%%%%%%%%%%%%%%%%%%%%%%%%%%%%%%%%%%%%%%%%%%%%%%%%%%%%%%%%%%%%

\subsection{Approximate recovery under Hamming error}
\label{SecHamming}

In the previous section, we analyzed performance in terms of exactly
recovering the $\toprank$-ranked subset.  Although exact recovery is
suitable for some applications (e.g., a setting with high stakes, in
which any single error has a large price), there are other settings in
which it may be acceptable to return a subset that is ``close'' to the
correct $\toprank$-ranked subset.  In this section, we analyze this
problem of approximate recovery when closeness is measured under the
Hamming error.  More precisely, for a given threshold $\hamthr \in [0,
  \toprank)$, suppose that our goal is to output a set
  $\toprank$-sized set $\tophat$ such that its Hamming distance to the
  set $\topstar$ of the true top $\toprank$ items, as defined in
  equation~\eqref{EqnDefnHammingSet}, is bounded as
\begin{align}
\label{eq:ham_requirement}
\dham(\tophat, \topstar) \leq 2\hamthr.
\end{align}
Our goal is to establish conditions under which it is possible (or
impossible) to return an estimate $\tophat$ satisfying the
bound~\eqref{eq:ham_requirement} with high probability.\footnote{The
  requirement $\hamthr < \toprank$ is sensible because if $\hamthr
  \geq \toprank$, the problem is trivial: any two $\toprank$-sized
  sets $\tophat$ and $\topstar$ satisfy the bound
  \mbox{$\dham(\tophat,\topstar) \leq 2 \toprank \leq 2\hamthr$.}}

As before, we use $(1),\ldots,(\numitems)$ to denote the permutation
of the $\numitems$ items in decreasing order of their scores.  With
this notation, the following quantity plays a central role in our
analysis:
\begin{subequations}
\begin{align}
\label{eq:defn_criticalKham}
\criticalKham \defn \score_{(\toprank - \hamthr)} - \score_{(\toprank
  + \hamthr + 1)}.
\end{align}
Observe that $\criticalKham$ is a generalization of the quantity
$\criticalK$ defined previously in equation~\eqref{eq:defn_criticalK};
more precisely, the quantity $\criticalK$ corresponds to
$\criticalKham$ with $\hamthr = 0$.  We then define a generalization
of the family $\FAMK{\KEYCON; \numitems, \pp, \numrepeat}$, namely
\begin{align}
\FAMKH{\KEYCON; \numitems, \pp, \numrepeat} & \defn \left \{ \Mmat \in
      [0,1]^{\numitems \times \numitems} \, \mid \, \Mmat + \Mmat^T =
      1 1^T, \mbox{ and } \criticalKham \geq \KEYCON \sqrt{\frac{ \log
          \numitems}{\numitems \pp \numrepeat}} \right \}.
\end{align}
\end{subequations}
As before, we frequently adopt the shorthand $\FAMKH{\KEYCON}$, with
the dependence on $(\numitems, \pp, \numrepeat)$ being understood
implicitly.
\begin{theorem}
\label{thm:topkham}
\begin{enumerate}
\item[(a)] For any $\KEYCON \geq 8$, the maximum pairwise win set
  $\toptilde$ satisfies
\begin{subequations}
\begin{align}
\label{eq:topkham_tail}
\sup_{\Mmat \in \FAMKH{\KEYCON}} \mprob_{\Mmat} \big[ \dham(\toptilde,
  \topstar) > 2 \hamthr \big] & \leq \frac{1}{\numitems^{14}}.
\end{align}
\item[(b)] Conversely, in the regime $\pp \geq \frac{\log \numitems}{2
  \numitems \numrepeat}$ and for given constants $\hamconexp,
  \hamconmul \in (0,1)$, suppose that $2\hamthr \leq \frac{1}{1 +
    \hamconmul} \min\{\numitems^{1 - \hamconexp}, \toprank, \numitems
  - \toprank \}$.  Then for any $\KEYCON \leq \frac{\sqrt{\hamconexp
      \hamconmul}}{14}$, any estimator $\tophat$ has error at least
\begin{align}
\sup_{\Mmat \in \FAMKH{\KEYCON}} \mprob_{\Mmat} \big[ \dham(\tophat ,
  \topstar) > 2\hamthr \big] & \geq \frac{1}{7},
\end{align}
for all $\numitems$ larger than a constant $c(\hamconexp,
\hamconmul)$.
\end{subequations}
\end{enumerate}
\end{theorem}

This result is similar to that of Theorem~\ref{thm:topk}, except that
the relaxation of the exact recovery condition allows for a less
constrained definition of the separation threshold $\criticalKham$.
As with Theorem~\ref{thm:topk}, the lower bound in part (b) applies
even if probability matrix $\Mmat$ is restricted to lie in a parametric
model (such as the BTL model), or the more general SST class.  The counting algorithm is thus optimal for estimation
under the relaxed Hamming metric as well.

Finally, it is worth making a few comments about the constants
appearing in these claims.  We can weaken the lower bound on
$\criticalK$ required in Theorem~\ref{thm:topkham}(a) at the expense
of a lower probability of success; for instance, if we instead require
that $\KEYCON \geq 4$, then the probability of error is guaranteed to
be at most $\numitems^{-2}$. Subsequently in the paper, we provide the
results of simulations with $\numitems = 500$ items and $\KEYCON = 4$.
On the other hand, in Theorem~\ref{thm:topkham}(b), if we impose the
stronger upper bound $\KEYCON = \order(1/\sqrt{\hamthr \log
  \numitems})$, then we can remove the condition $\hamthr \leq
\numitems^{1 - \hamconexp}$.

%%%%%%%%%%%%%%%%%%%%%%%%%%%%%%%%%%%%%%%%%%%%%%%%%%%%%%%%%%%%%%%%%%%%%

\subsection{An abstract form of $\toprank$-set recovery}
\label{SecGeneralReq}

In earlier sections, we investigated recovery of the top $\toprank$
items either exactly or under a Hamming error. Exact recovery may be
quite strict for certain applications, whereas the property of Hamming
error allowing for a few of the top $\toprank$ items to be replaced by
\emph{arbitrary} items may be undesirable. Indeed, many applications
have requirements that go beyond these metrics; for instance, see the
papers~\cite{ilyas2008survey,michel2005klee,babcock2003distributed,
  metwally2005efficient, kimelfeld2006finding, fagin2003optimal} and
references therein for some examples. In this section, we generalize
the notion of exact or Hamming-error recovery in order to accommodate
a fairly general class of requirements.

Both the exact and approximate Hamming recovery settings require the
estimator to output a set of $\toprank$ items that are either exactly
or approximately equal the true set of top $\toprank$ items. When is
the estimate deemed successful? One way to think about the problem is
as follows. The specified requirement of exact or approximate Hamming
recovery is associated to a set of $\toprank$-sized subsets of the
$\numitems$ possible ranks.  The estimator is deemed successful if the
true ranks of the chosen $\toprank$ items equals one of these
subsets. In our notion of generalized recovery, we refer to such sets
as \emph{allowed sets}. For example, in the case $\toprank = 3$, we
might say that the set $\{1,4,10\}$ is allowed, meaning that an output
consisting of the ``first'', ``fourth'' and ``tenth'' ranked items is
considered correct.

In more generality, let $\allallowedsets$ denote a family of
$\toprank$-sized subsets of $[\numitems]$, which we refer to as
\emph{family of allowed sets.}  Notice that any allowed set is defined
by the \emph{positions} of the items in the true ordering and not the
items themselves.\footnote{In case of two or more items with identical
  scores, the choice of any of these items is considered valid.}  Once
some true underlying ordering of the $\numitems$ items is fixed, each
element of the family $\allallowedsets$ then specifies a set of the
items themselves.  We use these two interpretations depending on the
context --- the definition in terms of positions to specify the
requirements, and the definition in terms of the items to evaluate an
estimator for a given underlying probability matrix $\Mmat$.

We let $\topdagger$ denote a $\toprank$-set estimate, meaning a
function that given a set of observations as input, returns a
$\toprank$-sized subset of $[\numitems]$ as output.
\begin{definition}[$\allallowedsets$-respecting estimators]
For any family $\allallowedsets$ of allowed sets, a $\toprank$-set
estimate $\topdagger$ \emph{respects} its structure if the set of
$\toprank$ positions of the items in $\topdagger$ belongs to the set
\mbox{family $\allallowedsets$.}
\end{definition}
\noindent Our goal is to determine conditions on the set family
$\allallowedsets$ under which there exist estimators $\topdagger$ that
respect its structure.  In order to illustrate this definition, let us
return to the examples treated thus far:
\begin{example}[Exact and approximate Hamming recovery]
\label{ExaVanilla}
The requirement of exact recovery of the top $\toprank$ items has
$\allallowedsets$ consisting of exactly one set, the set of the top
$\toprank$ positions $\allallowedsets = \{[\toprank]\}$. In the case
of recovery with a Hamming error at most $2 \hamthr$, the set
$\allallowedsets$ of all allowed sets consists all $\toprank$-sized
subsets of $[\numitems]$ that contain at least $(\toprank - \hamthr)$
positions in the top $\toprank$ positions.  For instance, in the case
$\hamthr = 1$, $\toprank = 2$ and $\numitems = 4$, we have
\begin{align*}
\allallowedsets & = \Big \{ \{1, 2 \}, \{1, 3\}, \{1, 4 \}, \{2, 3\},
\{2, 4\} \Big \}.
\end{align*}
\end{example}

\noindent Apart from these two requirements, there are several other
requirements for top-$\toprank$ recovery popular in the
literature~\cite{carmel2001static, fagin2003optimal,
  babcock2003distributed, michel2005klee, metwally2005efficient,
  kimelfeld2006finding, ilyas2008survey}.  Let us illustrate them with
another example:
\begin{example}
\label{ExOtherMetrics}
Let $\permstar: [\numitems] \rightarrow [\numitems]$ denote the true
underlying ordering of the $\numitems$ items. The following are four
popular requirements on the set $\topdagger$ for top-$\toprank$
identification, with respect to the true permutation $\permstar$, for
a pre-specified parameter $\epsothermetric \geq 0$.
\begin{subequations}
\begin{enumerate}[label = (\roman*)]
\item All items in the set $\topdagger$ must be contained contained
  within the top $(1 + \epsothermetric) \toprank$ entries:
\begin{align}
\max \limits_{i \in \topdagger} \permstar(i) \leq (1 +
\epsothermetric) \toprank.
\end{align}
\item The rank of any item in the set $\topdagger$ must lie within a
  multiplicative factor $(1 + \epsothermetric)$ of the rank of any
  item not in the set $\topdagger$:
\begin{align}
\max \limits_{i \in \topdagger} \permstar(i) \leq (1 +
\epsothermetric) \min \limits_{j \in [\numitems] \backslash
  \topdagger} \perm(j).
\end{align}
\item The rank of any item in the set $\topdagger$ must lie within an
  additive factor $\epsothermetric$ of the rank of any item not in the
  set $\topdagger$:
\begin{align}
\max \limits_{i \in \topdagger} \permstar(i) \leq \min \limits_{j \in
  [\numitems] \backslash \topdagger} \permstar(j) + \epsothermetric.
\end{align}
\item The sum of the ranks of the items in the set $\topdagger$ must
  be contained within a factor $(1 + \epsothermetric)$ of the sums of
  ranks of the top $\toprank$ entries:
\begin{align}
\sum \limits_{i \in \topdagger} \permstar(i) \leq (1 +
\epsothermetric) \half \toprank (\toprank + 1).
\end{align}
\end{enumerate}
\end{subequations}
Note that each of these requirements reduces to the exact recovery
requirement when $\epsothermetric = 0$.
Moreover, each of these requirements can be rephrased in terms of
families of allowed sets.  For instance, if we focus on requirement
(i), then any $\toprank$-sized subset of the top $(1 +
\epsothermetric) \toprank$ positions is an allowable set.
\end{example}

In this paper, we derive conditions that govern $\toprank$-set
recovery for allowable set systems that satisfy a natural
``monotonicity'' condition. Informally, the monotonicity condition
requires that the set of $\toprank$ items resulting from replacing an
item in an allowed set with a higher ranked item must also be an
allowed set.  More precisely, for any set $\{
\alloweditem_1,\ldots,\alloweditem_\toprank \} \subseteq [\numitems]$,
let $\monotone(\{\alloweditem_1,\ldots,\alloweditem_\toprank\})
\subseteq 2^{[\numitems]}$ be the set defined by all of its monotone
transformations---that is
\begin{align*}
\monotone(\{\alloweditem_1,\ldots,\alloweditem_\toprank\} ) \defn \Big
\{ \{\alloweditem'_1, \ldots, \alloweditem'_\toprank\} \subseteq
   [\numitems] \mid \alloweditem'_\varalloweditem \leq
   \alloweditem_\varalloweditem \mbox{ for every } \varalloweditem \in
               [\toprank] \Big \}.
\end{align*}
Using this notation, we have the following:
\begin{definition}[Monotonic set systems]
The set $\allallowedsets$ of allowed sets is a \emph{monotonic set
  system} if
\begin{align}
\label{EqnDefnMonotonicSetSystem}
\monotone(\allowedset) \subseteq \allallowedsets \quad \mbox{for every
} \allowedset \in \allallowedsets.
\end{align}
\end{definition}
\noindent One can verify that
condition~\eqref{EqnDefnMonotonicSetSystem} is satisfied by the
settings of exact and Hamming-error recovery, as discussed in
Example~\ref{ExaVanilla}. The condition is also satisfied by all four
requirements discussed in Example~\ref{ExOtherMetrics}.

The following theorem establishes conditions under which one can (or
cannot) produce an estimator that respects an allowable set
requirement.  In order to state it, recall the score $\score_i \defn
\frac{1}{\numitems} \sum_{j = 1}^{\numitems} \compare{i}{j}$, as
previously defined in equation~\eqref{eq:defn_scores} for each $i \in
[\numitems]$.  For notational convenience, we also define $\score_i
\defn - \infty$ for every $i > \numitems$.  Consider any monotonic
family of allowed sets $\allallowedsets$, and for some integer
$\numallowedsets \geq 1$, let
$\allowedset^1,\ldots,\allowedset^\numallowedsets \in \allallowedsets$
such that $\allallowedsets = \cuplimits \limits_{\varallowedset \in
  [\numallowedsets]} \monotone(\allowedset^\varallowedset)$. For every
$\varallowedset \in [\numallowedsets]$, let
$\alloweditem_1^\varallowedset < \cdots <
\alloweditem_\toprank^\varallowedset$ denote the entries of
$\allowedset^\varallowedset$.  We then define the critical threshold
based on the scores:
\begin{align}
\label{eq:defn_criticalS}
\criticalS \defn \max \limits_{\varallowedset \in [\numallowedsets]}
\min \limits_{\varalloweditem \in [\toprank]}
(\score_{(\varalloweditem)} - \score_{(\toprank +
  \alloweditem_\varalloweditem^\varallowedset - \varalloweditem +
  1)}).
\end{align}

The term $\criticalS$ is a further generalization of the quantities
$\criticalK$ and $\criticalKham$ defined in earlier sections. We also
define a generalization $\FAMSET{\cdot}$ of the families
$\FAMK{\cdot}$ and $\FAMK{\cdot}$ as
\begin{align}
\FAMSET{\KEYCON; \numitems, \pp, \numrepeat} & \defn \left \{ \Mmat
\in [0,1]^{\numitems \times \numitems} \, \mid \, \Mmat + \Mmat^T = 1
1^T \mbox{ and } \criticalS \geq \KEYCON \sqrt{\frac{ \log
    \numitems}{\numitems \pp \numrepeat}} \right \}.
\end{align}
As before, we use the shorthand $\FAMSET{\KEYCON}$, with the
dependence on $(\numitems, \pp, \numrepeat)$ being understood
implicitly.
\begin{theorem}
\label{ThmGeneralReq}
Consider any allowable set requirement specified by a monotonic set
class $\allallowedsets$.
\begin{enumerate}
\item[(a)] For any $\KEYCON \geq 8$, the maximum pairwise win set
  $\toptilde$ satisfies
\begin{align*}
\sup_{\Mmat \in \FAMSET{\KEYCON}} \mprob_{\Mmat} \big[ \toptilde
  \notin \allallowedsets \big] & \leq \frac{1}{\numitems^{13}}.
\end{align*}
\item[(b)] Conversely, in the regime $\pp \geq \frac{\log \numitems}{2
  \numitems \numrepeat}$, and for given constants $\genconexp \in
  (0,1), \genconmul \in (\frac{3}{4},1]$, suppose that
$\max_{\varallowedset \in [\numallowedsets]} \alloweditem_{\lceil
  \genconmul \toprank \rceil}^{\varallowedset} \leq
\frac{\numitems}{2}$ and $8 ( 1 - \genconmul) \toprank \leq
\numitems^{1 - \genconexp }$. Then for any $\KEYCON$ smaller than a
constant $\UUP(\genconexp, \genconmul)>0$, any estimator $\tophat$ has
error at least
\begin{align}
\sup_{\Mmat \in \FAMSET{\KEYCON}} \mprob_{\Mmat} \big[\toptilde \notin
  \allallowedsets \big] & \geq \frac{1}{15},
\end{align}
for all $\numitems$ larger than a constant $\UNUM(\genconexp,
\genconmul)$.
\end{enumerate}
\end{theorem}

A few remarks on the lower bound are in order.  First, the lower bound
continues to hold even if the probability matrix $\Mmat$ is restricted to follow a parametric model such as BTL or restricted to lie in the SST class. Second, in terms of the threshold for
$\KEYCON$, the lower bound holds with $\UUP(\genconexp, \genconmul) =
\frac{1}{15} \sqrt{\genconexp \min\big\{\frac{1}{4(1-\genconmul)-1},
  \half} \big\}$.  Third, it is worth noting that one must necessarily
impose some conditions for the lower bound, along the lines of those
required in Theorem~\ref{ThmGeneralReq}(b) for the allowable sets to
be ``interesting'' enough.

As a concrete illustration, consider the requirement defined by the
parameters $\varallowedset = 1$, $\toprank = 1$ and $\allallowedsets =
\monotone(\{\numitems - \sqrt{\numitems}\})$. For $\genconexp =
\genconmul = \frac{9}{10}$, this requirement satisfies the condition
$8(1 - \genconmul) \toprank \leq \numitems^{1 - \genconexp}$ but
violates the condition $\alloweditem_{\lceil\genconmul \toprank
  \rceil} \leq \frac{\numitems}{2}$. Now, a selection of $\toprank=1$
item made uniformly at random (independent of the data) satisfies this
allowable set requirement with probability $1 -
\frac{1}{\sqrt{\numitems}}$. Given the success of such a random
selection algorithm in this parameter regime, we see that the lower
bounds therefore cannot be universal, but must require some conditions
on the allowable sets.

%%%%%%%%%%%%%%%%%%%%%%%%%%%%%%%%%%%%%%%%%%%%%%%%%%%%%%%%%%%%

\section{Simulations and experiments}
\label{SecSimulations}

In this section, we empirically evaluate the performance of the
counting algorithm and compare it with the Spectral MLE algorithm via
simulations on synthetic data, as well as experiments using datasets
from the Amazon Mechanical Turk crowdsourcing platform.

%%%%%%%%%%%%%%%%%%%%%%%%%%%%%%%%%%%%%%%%%%%%%%%%%%%%%%%%%%%%%%%%%%%%%%

\subsection{Simulated data}

\begin{figure*}
\centering \includegraphics[width = .9\textwidth]{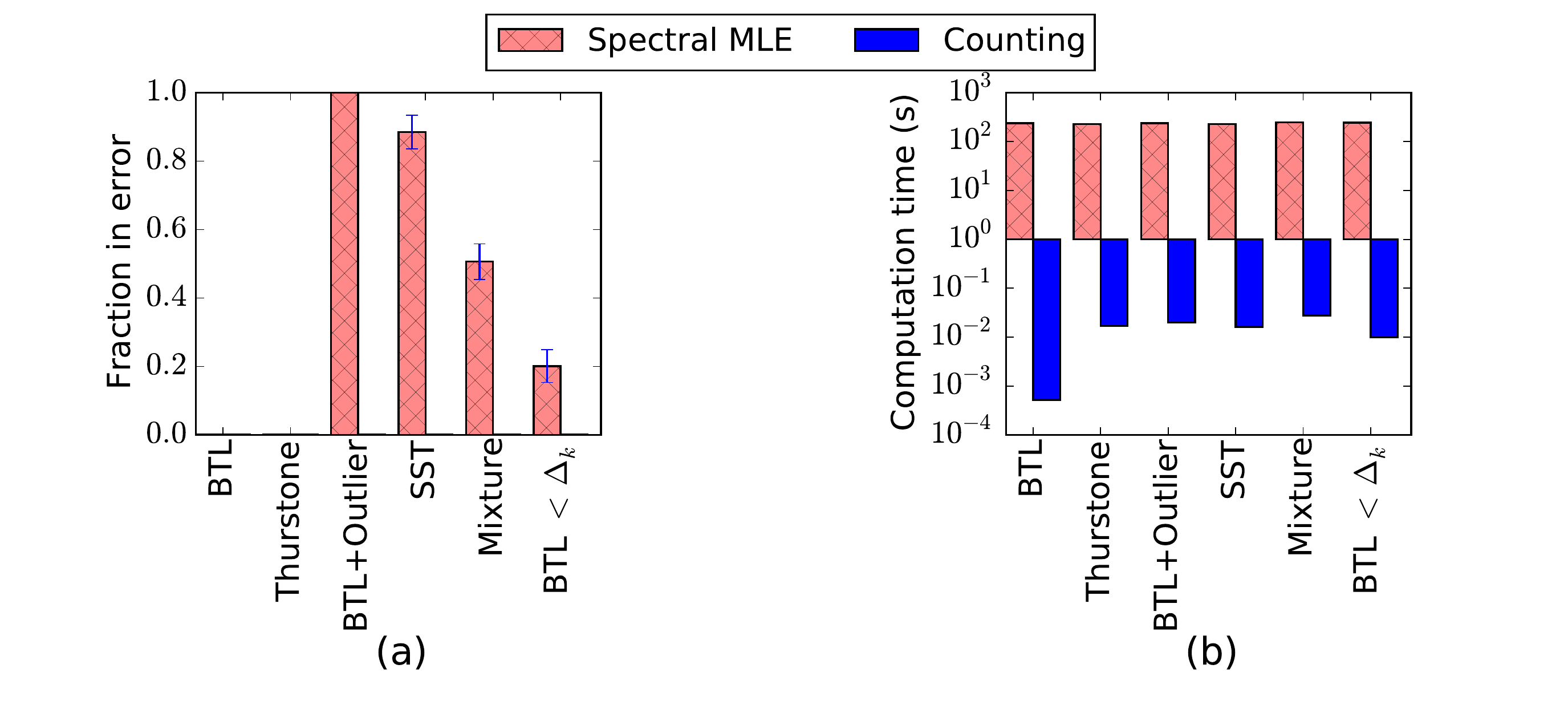}
\caption{Simulation results comparing Spectral MLE and the counting
  algorithm in terms of error rates for exact recovery of the top
  $\toprank$ items, and computation time.  (a) Histogram of fraction
  of instances where the algorithm failed to recover the $\toprank$
  items correctly, with each bar being the average value across $50$
  trials.  The counting algorithm has 0\% error across all problems,
  while the spectral MLE is accurate for parametric models (BTL,
  Thurstone), but increasingly inaccurate for other models.  (b)
  Histogram plots of the maximum computation time taken by the
  counting algorithm and the minimum computation time taken by
  Spectral MLE across all trials.  Even though this maximum-to-minimum
  comparison is unfair to the counting algorithm, it involves five or
  more orders of magnitude less computation.}
\label{fig:simulations}
\end{figure*}

We begin with simulations using synthetically generated data with
$\numitems = 500$ items and observation probability $\pp = 1$, and
with pairwise comparison models ranging over six possible types.
Panel (a) in Figure~\ref{fig:simulations} provides a histogram plot of
the associated error rates (with a bar for each one of these six
models) in recovering the $\toprank = \numitems/4 = 125$ items for the
counting algorithm versus the Spectral MLE algorithm.  Each bar
corresponds to the average over $50$ trials.  Panel (b) compares the
CPU times of the two algorithms. The value of $\KEYCON$ (and in turn, the value of $\numrepeat$) in the first five models is as derived in Section~\ref{SecExact}. In more detail, the six model types
are given by:
\begin{enumerate}[leftmargin=*]

\item[(I)] \emph{Bradley-Terry-Luce (BTL) model:} Recall that the
  theoretical guarantees for the Spectral MLE
  algorithm~\cite{chen2015spectral} are applicable to data that is
  generated from the BTL model~\eqref{eq:defn_BTL}, and as guaranteed,
  the Spectral MLE algorithm gives a $100\%$ accuracy under this
  model. The counting algorithm also obtains a $100\%$ accuracy, but
  importantly, the counting algorithm requires a computational time
  that is five orders of magnitude lower than that of Spectral MLE.
\item[(II)] \emph{Thurstone model:} The Thurstone
  model~\cite{thurstone1927law} is another parametric model, with the
  function $F$ in equation~\eqref{eq:defn_parametric} set as the
 cumulative distribution function of the standard Gaussian distribution. Both Spectral MLE and
  the counting algorithm gave $100\%$ accuracy under this model.

\item[(III)] \emph{BTL model with one (non-transitive) outlier:} This
  model is identical to BTL, with one modification. Comparisons among
  $(\numitems - 1)$ of the items follow the BTL model as before, but
  the remaining item always beats the first $\frac{\numitems}{4}$
  items and always loses to each of the other items. We see that the
  counting algorithm continues to achieve an accuracy of $100\%$ as
  guaranteed by Theorem~\ref{thm:topk}. The departure from the BTL
  model however prevents the Spectral MLE algorithm from identifying
  the top $\toprank$ items.

\item[(IV)] \emph{Strong stochastic transitivity (SST) model:} We
  simulate the ``independent diagonals'' construction
  of~\cite{shah2015stochastically} in the SST class.  Spectral MLE is
  often unsuccessful in recovering the top $\toprank$ items, while the
  counting algorithm always succeeds.

\item[(V)] \emph{Mixture of BTL models:} Consider two sets of people
  with opposing preferences.  The first set of people have a certain
  ordering of the items in their mind and their preferences follow a
  BTL model under this ordering. The second set of people have the
  opposite ordering, and their preferences also follow a BTL model
  under this opposite ordering. The overall preference probabilities
  is a mixture between these two sets of people. In the simulations,
  we observe that the counting algorithm is always successful while
  the Spectral MLE method often fails.

\item[(VI)] \emph{BTL with violation of separation condition:} We
  simulate the BTL model, but with a choice of parameter $\numrepeat$
  small enough that the value of $\KEYCON$ is about one-tenth of
  its recommended value in Section~\ref{SecExact}. We observe that
  the counting algorithm incurs lower errors than the Spectral MLE
  algorithm, thereby demonstrating its robustness.

\end{enumerate}

To summarize, the performance of the two algorithms can be contrasted
in the following way.  When our stated lower bounds on $\KEYCON$ are
satisfied, then consistent with our theoretical claims, the Copeland
counting algorithm succeeds irrespective of the form of the pairwise
probability distributions. The Spectral MLE algorithm performs well
when the pairwise comparison probabilities are faithful to parametric
models, but is often unsuccessful otherwise. Even when the condition
on $\KEYCON$ is violated, the performance of the counting algorithm
remains superior to that of the Spectral MLE.\footnote{Note that part
  (b) of Theorem~\ref{thm:topk} is a minimax converse meaning that it
  appeals to the worst case scenario.} In terms of computational
complexity, for every instance we simulated, the counting algorithm
took several orders of magnitude less time as compared to Spectral
MLE.

%%%%%%%%%%%%%%%%%%%%%%%%%%%%%%%%%%%%%%%%%%%%%%%%%%%%%

\subsection{Experiments on data from Amazon Mechanical Turk}

In this section, we describe experiments on real world datasets
collected from the Amazon Mechanical Turk (\url{mturk.com}) commercial
crowdsourcing platform.

%%%%%%%%%%%%%%%%%%

\subsubsection{Data}

In order to evaluate the accuracy of the algorithms under
consideration, we require datasets consisting of pairwise comparisons
in which the questions can be associated with an objective and
verifiable ground truth. To this end, we used the ``cardinal versus
ordinal'' dataset from our past work~\cite{shah2015estimation}; three
of the experiments performed in that paper are suitable for the
evaluations here---namely, ones in which each question has a ground
truth, and the pairs of items are chosen uniformly at random.  The
three experiments tested the workers' general knowledge, audio, and
visual understanding, and the respective tasks involved: (i)
identifying the pair of cities with a greater geographical distance,
(ii) identifying the higher frequency key of a piano, and (iii)
identifying spelling mistakes in a paragraph of text. The number of
items $\numitems$ in the three experiments were $16$, $10$ and $8$
respectively. The total number of pairwise comparisons were $408$,
$265$ and $184$ respectively. The fraction of pairwise comparisons
whose outcomes were incorrect (as compared to the ground truth) in the
raw data are $17\%$, $20\%$ and $40\%$ respectively.

%%%%%%%%%%%%%%%%%%%%%%%%%%%%%%%

\subsubsection{Results}

We compared the performance of the counting algorithm with that of the
Spectral MLE algorithm. For each value of a ``subsampling
probability'' $q \in \{0.1,0.2,\ldots,1.0\}$, we subsampled a fraction
$q$ of the data and executed both algorithms on this subsampled
data. We evaluated the performance of the algorithms on their ability
to recover the top $\toprank = \lceil \frac{\numitems}{4} \rceil$
items under the Hamming error metric.

\begin{figure}
\centering \includegraphics[width = \textwidth]{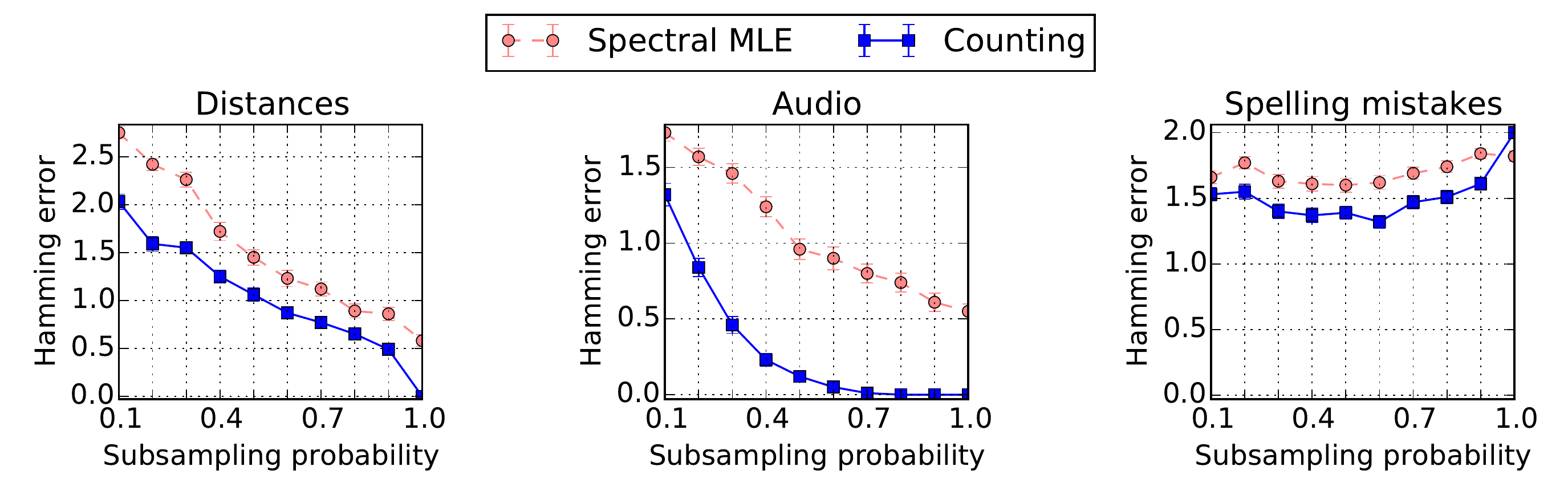}
\caption{Evaluation of Spectral MLE and the counting algorithm on
  three datasets from Amazon Mechanical Turk in terms of the error
  rates for top $\toprank$-subset recovery. The three panels plot the
  Hamming error when recovering the top $\toprank$ items in the three
  datasets when a $q^{th}$ fraction of the total data is used, for
  various values of subsampling probability $q \in (0,1]$. The
counting algorithm consistently outperforms the Spectral MLE
algorithm.}
\label{fig:mturk_hamming}
\end{figure}

Figure~\ref{fig:mturk_hamming} shows the results of the
experiments. Each point in the plots is an average across $100$
trials. Observe that the counting algorithm consistently outperforms
Spectral MLE. (We think that the erratic fluctuations in the spelling
mistakes data are a consequence of a high noise and a relatively small
problem size.) Moreover, the Spectral MLE algorithm required about $5$
orders of magnitude more computation time (not
shown in the figure) as compared to counting. Thus the counting algorithm performs well on
simulated as well as real data. It outperforms Spectral MLE not only
when the number of items is large (as in the simulations) but also
when the problem sizes are small as seen in these experiments.

%%%%%%%%%%%%%%%%%%%%%%%%%%%%%%%%%%%%%%%%%%%%%%%%%%%%%%%%%%%%%%%%%%%%%%

\section{Proofs}
\label{SecProofs}

We now turn to the proofs of our main results.  We continue to use the
notation $[i]$ to denote the set $\{1,\ldots,i\}$ for any integer $i
\geq 1$.  We ignore floor and ceiling conditions unless critical to
the proof.

Our lower bounds are based on a standard form of Fano's
inequality~\cite{cover2012elements,Tsybakovbook} for lower bounding
the probability of error in an $\mary$-ary hypothesis testing problem.
We state a version here for future reference.  For some integer $\mary
\geq 2$, fix some collection of distributions $\{\mprob^1, \ldots,
\mprob^\mary\}$.  Suppose that we observe a random variable $Y$ that is
obtained by first sampling an index $A$ uniformly at random from
$[\mary] = \{1, \ldots, \mary\}$, and then drawing $Y \sim \mprob^A$.
(As a result, the variable $Y$ is marginally distributed according to
the mixture distribution \mbox{$\Pbar = \frac{1}{\mary}
  \sum_{a=1}^\mary \mprob^a$.)}  Given the observation $Y$, our goal
is to ``decode'' the value of $A$, corresponding to the index of the
underlying mixture component.  Using $\mathcal{Y}$ to denote the
sample space associated with the observation $Y$, Fano's inequality
asserts that any test function $\PLAINTEST: \mathcal{Y} \rightarrow
[\mary]$ for this problem has error probability lower bounded as
\begin{align*}
\mprob[\TESTY \neq A] & \geq 1 - \frac{I(Y; A) + \log 2}{\log \mary},
\end{align*}
where $I(Y; A)$ denotes the mutual information between $Y$ and $A$.
A standard convexity argument for the mutual information yields the
weaker bound
\begin{align}
\label{EqnWeakFano}
\mprob[\TESTY \neq A] & \geq 1 - \frac{\max \limits_{a, b \in [\mary]}
  \kl{\mprob^a}{\mprob^b} + \log 2}{\log \mary},
\end{align}
We make use of this weakened form of Fano's inequality in several proofs.

%%%%%%%%%%%%%%%%%%%%%%%%%%%%%%%%%%%%%%%%%%%%%%%%%%%%%%%%%%%%%%%%%%%%

\subsection{Proof of Theorem~\ref{thm:topk}}

We begin with the proof of Theorem~\ref{thm:topk}, dividing our
argument into two parts.

%%%%%%%%%%%%%%%%%%%%%%%%%%%%%%%%%%%%%%%%%%%%%%%%%%%%%%%%%%%%%%%%%%%%%%%%%%

\subsubsection{Proof of part (a)}

For any pair of items $(i,j)$, let us encode the outcomes of the
$\numrepeat$ trials by an i.i.d. sequence \mbox{${V}_{ij}^{(\repvar)}
  = [\itemrv_{ij}^{(\repvar)}~~~\itemrv_{ji}^{(\repvar)}]^T$} of
random vectors, indexed by $\repvar \in [\numrepeat]$.  Each random
vector follows the distribution
\begin{align*}
\mprob \big[x_{ij}^{(\repvar)}, \; x_{ji}^{(\repvar)} \big] &
= \begin{cases}
1 - \pp & \mbox{if $(x_{ij}^{(\repvar)}, \; x_{ji}^{(\repvar)}) =
  (0,0)$} \\
\pp \Mmat_{ij} & \mbox{if $(x_{ij}^{(\repvar)}, \; x_{ji}^{(\repvar)})
  = (1,0)$} \\
\pp (1 - \Mmat_{ij} ) & \mbox{if $(x_{ij}^{(\repvar)}, \;
  x_{ji}^{(\repvar)}) = (0,1)$} \\
0 & \mbox{otherwise.}
\end{cases}
\end{align*}
With this encoding, the variable $W_a \defn \sum_{\ell \in
  [\numrepeat]} \sum_{z \in [\numobs] \backslash \{a\}} X_{a
  j}^{(\numrepeat)}$ encodes the number of wins for \mbox{item $a$.}

Consider any item $\itemup \in \topstar$ which ranks among the top
$\toprank$ in the true underlying ordering, and any item $\itemdown
\in [\numitems] \backslash \topstar$ which ranks outside the top
$\toprank$. We claim that with high probability, item $\itemup$ will
win more pairwise comparisons than item $\itemdown$. More precisely,
let $\event_{\itemdown \itemup}$ denote the event that item
$\itemdown$ wins at least as many pairwise comparisons than
$\itemup$. We claim that
\begin{align}
\label{EqnBernsteinConclusion}
\mprob(\event_{ \itemdown \itemup}) & \stackrel{(i)}{\leq} \exp \left(
- \frac{ \tfrac{1}{2} (\numrepeat \pp \numitems \criticalK)^2 }{
  \numrepeat \pp \numitems (2 - \criticalK) + \tfrac{2}{3} \numrepeat
  \pp \numitems \criticalK } \right) \; \stackrel{(ii)}{\leq} \;
\frac{1}{\numitems^{16}}.
\end{align}
Given this bound, the probability that the counting algorithm will
rank item $\itemdown$ above $\itemup$ is no more than
$\numitems^{-16}$. Applying the union bound over all pairs of items
$\itemup \in \topstar$ and $\itemdown \in [\numitems] \backslash
\topstar$ yields $\mprob \big[ \toptilde \neq \topstar \big] \leq
\numobs^{-14}$ as claimed.

We note that inequality (ii) in
equation~\eqref{EqnBernsteinConclusion} follows from inequality (i)
combined with the condition on $\criticalK$ that arises by setting $\KEYCON \geq 8$ as assumed in the hypothesis of the
theorem.  Thus, it remains to prove \mbox{inequality (i)} in
equation~\eqref{EqnBernsteinConclusion}.  By definition of
$\event_{\itemdown \itemup}$, we have
\begin{align}
\mprob (\event_{\itemdown \itemup}) & = \mprob \Big(
\underbrace{\sum_{\repvar \in [\numrepeat]} \sum_{z \in [\numitems]
    \backslash \{\itemdown\}} \itemrv_{\itemdown
    z}^{(\repvar)}}_{W_\itemdown} - \underbrace{\sum_{\repvar \in
    [\numrepeat]} \sum_{z \in [\numitems] \backslash \{\itemup\}}
  \itemrv_{\itemup z}^{(\repvar)}}_{W_a} \geq 0 \Big).
\end{align}
It is convenient to recenter the random variables. For every $\repvar
\in [\numrepeat]$ and $z \in [\numitems] \backslash \{\itemup,
\itemdown\}$, define the zero-mean random variables
\begin{align*}
\itemrvcent_{\itemup z}^{(\repvar)} = \itemrv_{\itemup z}^{(\repvar)}
- \Exs[\itemrv_{\itemup z}^{(\repvar)}] = \itemrv_{\itemup
  z}^{(\repvar)} - \pp \compare{\itemup}{z} ~~ \mbox{and} ~~
\itemrvcent_{\itemdown z}^{(\repvar)} = \itemrv_{\itemdown
  z}^{(\repvar)} - \Exs[\itemrv_{\itemdown z}^{(\repvar)}] =
\itemrv_{\itemdown z}^{(\repvar)} - \pp \compare{\itemdown}{z}.
\end{align*} 
Also, let
\begin{align*}
\itemrvcent_{\itemup \itemdown}^{(\repvar)} &= (\itemrv_{\itemup
  \itemdown}^{(\repvar)} - \itemrv_{\itemdown \itemup}^{(\repvar)}) -
\Exs[\itemrv_{\itemup \itemdown}^{(\repvar)} - \itemrv_{\itemdown
    \itemup}^{(\repvar)}] \; = \; (\itemrv_{\itemup
  \itemdown}^{(\repvar)} - \itemrv_{\itemdown \itemup}^{(\repvar)}) -
(\pp \compare{\itemup}{\itemdown} - \pp \compare{\itemdown}{
  \itemup}).
\end{align*}
We then have
\begin{align*}
\mprob(\event_{\itemdown \itemup}) & = \mprob \Bigg( \sum_{\ell \in
  [\numrepeat]} \Big( \sum_{z \in [\numitems] \!\! \backslash
  \{\itemup, \itemdown\}} \itemrvcent_{\itemdown z}^{(\repvar)} -
\sum_{z \in [\numitems] \backslash \{\itemup, \itemdown\}} \!\!
\itemrvcent_{\itemup z}^{(\repvar)} - \itemrvcent_{\itemup
  \itemdown}^{(\repvar)} \Big) \geq \numrepeat \pp \sum_{z \in
  [\numitems]} \Big( \compare{\itemup}{z} - \compare{\itemdown}{z}
\Big) \Bigg).
\end{align*}
Since $\itemup \in \topstar$ and $\itemdown \in [\numitems] \backslash
\topstar$, from the definition of $\criticalK$, we have $\numitems
\criticalK \leq \sum \limits_{z \in [\numitems]} \left( \Mmat_{\itemup
  z} - \Mmat_{\itemdown z} \right)$, and consequently
\begin{align}
\mprob \left( \event_{\itemdown \itemup} \right) & \leq \mprob \left(
\sum_{\ell \in [\numrepeat]} \Big( \sum_{z \in [\numitems] \!\!
  \backslash \{\itemup, \itemdown\}} \itemrvcent_{\itemdown
  z}^{(\repvar)} - \sum_{z \in [\numitems] \backslash \{\itemup,
  \itemdown\}} \!\! \itemrvcent_{\itemup z}^{(\repvar)} -
\itemrvcent_{\itemup \itemdown}^{(\repvar)} \Big) \geq \numrepeat \pp
\numitems \criticalK \right).
\label{eq:proof_upper_3}
\end{align}

By construction, all the random variables in the above inequality are
zero-mean, mutually independent, and bounded in absolute value by $2$.
These properties alone would allow us to obtain a tail bound by
Hoeffding's inequality; however, in order to obtain the stated
result~\eqref{EqnBernsteinConclusion}, we need the more refined result
afforded by Bernstein's inequality
(e.g.,~\cite{boucheron2013concentration}).  In order to derive a bound
of Bernstein type, the only remaining step is to bound the second
moments of the random variables at hand. Some straightforward
calculations yield
\begin{align*}
\Exs [ ( -\itemrvcent_{\itemup z}^{(\repvar)})^2] \leq \pp
\compare{\itemup}{z}, & \qquad \Exs [ ( \itemrvcent_{\itemdown
    z}^{(\repvar)})^2] \leq \pp \compare{\itemdown}{z}, \quad
\mbox{and} \quad
\Exs [ ( \itemrvcent_{\itemup \itemdown}^{(\repvar)})^2] \leq \pp
\compare{\itemup}{\itemdown} + \pp \compare{\itemdown}{\itemup}.
\end{align*}
It follows that
\begin{align*}
\sum_{z \in [\numitems] \backslash \{\itemup,\itemdown\} } \Exs [ (
  -\itemrvcent_{\itemup z}^{(\repvar)})^2] + & \sum_{z \in [\numitems]
  \backslash \{\itemup,\itemdown\} } \Exs [ ( \itemrvcent_{\itemdown
    z}^{(\repvar)})^2] + \Exs[ ( \itemrvcent_{\itemup
    \itemdown}^{(\repvar)})^2 ] \nonumber \\
&\leq \pp \left( \sum_{z\in [\numitems] \backslash \{\itemup,
  \itemdown\}} ( \compare{\itemup}{z} + \compare{\itemdown}{z} ) +
\compare{\itemup}{\itemdown} + \compare{\itemdown}{\itemup} \right)
\nonumber \\
& \stackrel{(iii)}{\leq} \pp \left( 2\sum_{z \in [\numitems]}
\compare{a}{z} - \numitems \criticalK \right)\\
& \stackrel{(iv)}{<} \pp \numitems (2 - \criticalK),
\end{align*}
where the inequality (iii) follows from the definition of
$\criticalK$, and step (iv) follows because $\compare{\itemup}{z} \leq
1$ for every $z$ and $\compare{\itemup}{\itemup} =
\frac{1}{2}$. Applying the Bernstein inequality now yields the stated
bound~\eqref{EqnBernsteinConclusion}(i).

%%%%%%%%%%%%%%%%%%%%%%%%%%%%%%%%%%%%%%%%%%%%%%%%%%%%%%%%%%%%%%%%%%%%%%%%%%%%

\subsubsection{Proof of part (b)}

The symmetry of the problem allows us to assume, without loss of
generality, that $\toprank \leq \frac{\numitems}{2}$.  We prove a
lower bound by first constructing a ensemble of $\numitems - \toprank
+ 1$ different problems, and considering the problem of distinguishing
between them.  For each $a \in \{ \toprank-1, \toprank, \ldots,
\numobs\}$, let us define the $\toprank$-sized subset $\Sstar[a] \defn
\{1, \ldots, \toprank -1 \} \cup \{a \}$, and the associated matrix of
pairwise probabilities
\begin{align*}
\Mmat^a_{ij} & \defn
\begin{cases} 
\frac{1}{2} & \mbox{if $i, j \in \Sstar[a]$, or $i,j \notin
  \Sstar[a]$} \\
\frac{1}{2} + \delta & \mbox{if $i \in \Sstar[a]$ and $j \notin
  \Sstar[a]$} \\
\frac{1}{2} - \delta & \mbox{if $i \notin \Sstar[a]$ and $j \in
  \Sstar[a]$,} \\
\end{cases}
\end{align*}
where $\delta \in (0, \frac{1}{2})$ is a parameter to be chosen.  We
use $\mprob^a$ to denote probabilities taken under pairwise
comparisons drawn according to the model $\Mmat^a$.

One can verify that the construction above falls in the intersection
of parametric models and the SST model. In the parametric case, this
construction amounts to having the parameters associated to every item
in $\topstar$ to have the same value, and those associated to every
item in $[\numitems] \backslash \topstar$ to have the same value. Also
observe that for every such distribution $\mprob^a$, the associated
$\toprank$-separation threshold $\criticalK = \delta$.

Any given set of observations can be described by the collection of
random variables \mbox{$Y = \{ Y_{ij}^{(\repvar)}, j > i \in
  [\numitems], \repvar \in [\numrepeat] \}$.}  When the true
underlying model is $\mprob^a$, the random variable
$Y_{ij}^{(\repvar)}$ follows the distribution
\begin{align*}
Y_{ij}^{(\repvar)} =
\begin{cases}
0 & \mbox{with probability } 1 - \pp \\ i & \mbox{with probability }
\pp \Mmat^a_{i j} \\ j & \mbox{with probability } \pp (1-
\Mmat^a_{i  j} ).
\end{cases}
\end{align*}
The random variables $\{ Y_{ij}^{(\repvar)} \}_{i,j \in [\numitems], i
  < j, \repvar \in [\numrepeat] }$ are mutually independent, and the distribution $\mprob^a$ is a product distribution across pairs $\{i > j\}$ and
repetitions $\repvar \in [\numrepeat]$.

Let $A \in \{ \toprank, \ldots, \numobs \}$ follow a uniform
distribution over the index set, and suppose that given $A = a$, our
observations $\obs$ has components drawn according to the model
$\mprob^a$.  Consequently, the marginal distribution of $Y$ is the
mixture distribution $\frac{1}{\mary} \sum_{a =1}^\mary \mprob^a$ over all $\mary
= \numobs - \toprank + 1$ models.  Based on observing $\obs$, our goal is
to recover the correct index $A = a$ of the underlying model, which is
equivalent to recovering the planted subset $\Sstar[a]$. We use the
Fano bound~\eqref{EqnWeakFano} to lower bound the error bound
associated with any test for this problem. In order to apply Fano's
inequality, the following result provides control over the
Kullback-Leibler divergence between any pair of probabilities
involved.
\begin{lemma}
\label{lem:KL}
For any distinct pair $a,b \in \{\toprank, \ldots,\numitems\}$, we
have
\begin{align}
\kl{\mprob^a}{\mprob^b} \leq \frac{ 2\numitems \pp
  \numrepeat}{\frac{1}{4 \delta^2} - 1}.
\end{align}
\end{lemma}
\noindent See the end of this section for the proof of this claim.

Given this bound on the Kullback-Leibler divergence, Fano's
inequality~\eqref{EqnWeakFano} implies that any estimator $\PLAINTEST$
of $A$ has error probability lower bounded as
\begin{align*}
\mprob[ \TESTY \neq A] & \geq 1 - \frac{ \frac{ 2\numitems \pp
    \numrepeat}{\frac{1}{4 \delta^2} - 1} + \log 2}{\log (\numitems -
  \toprank+1)} \; \geq \; \frac{1}{7}.
\end{align*}
Here the final inequality holds whenever $\delta \leq \frac{1}{7}
\sqrt{\frac{\log \numitems}{\numitems \pp \numrepeat}}$, $\pp \geq
\frac{\log \numitems}{2 \numitems \numrepeat}$, $\numitems \geq 7$ and
$\toprank \leq \frac{\numitems}{2}$. The condition $\pp \geq
\frac{\log \numitems}{2 \numitems \numrepeat}$ also ensures that
$\delta < \frac{1}{2}$ thereby ensuring that our construction is
valid. It only remains to prove Lemma~\ref{lem:KL}.

%%%%%%%%%%%%%%%%%%%%%%%%%%%%%%%%%%%%%%%%%%%%%%%%%%%%%%%%%%%%%%%%%%%%%%%%

\subsubsection{Proof of Lemma~\ref{lem:KL}} 
\label{SecProofLemKL}
Since the distributions $\mprob^a$ and $\mprob^b$ are formed by
components that are independent across edges $i > j$ and repetitions
$\repvar \in [\numrepeat]$, we have
\begin{align*}
\kl{ \mprob^a}{\mprob^b} = \sum_{\repvar \in [\numrepeat]} \sum_{1
  \leq i < j \leq \numitems} \kl{ \mprob^a( \itemrv_{ij}^{(\repvar)}
  )}{ \mprob^b (\itemrv_{ij}^{(\repvar)})} \; = \; \numrepeat \sum_{1
  \leq i < j \leq \numitems} \kl{ \mprob^a( \itemrv_{ij}^{(1)} )}{
  \mprob^b (\itemrv_{ij}^{(1)})},
\end{align*}
where the second equality follows since the $\numrepeat$ trials are
all independent and identically distributed.  

We now evaluate each individual term in right hand side of the above
equation. Consider any $i, j \in [\numitems]$.  We divide our analysis
into three disjoint cases:\\
{\underline{Case I}: Suppose that $i,j \in [\numitems] \backslash
  \{a,b\}$.}  The distribution of $\itemrv_{ij}^{(1)} $ is
identical across the distributions $\mprob^a$ and $\mprob^b$. As a
result, we find that
\begin{align*}
\kl{ \mprob^a(\itemrv_{ij}^{(1)}) }{ \mprob^b
  (\itemrv_{ij}^{(1)} )} = 0.
\end{align*}
\\
{\underline{Case II}: Suppose that $i = a,\ j \in [\numitems]
  \backslash \{a,b\}$ or $i = b,\ j \in [\numitems] \backslash \{a,
  b\}$.} We then have
\begin{align*}
\kl{ \mprob^a(\itemrv_{ij}^{(1)}) }{ \mprob^b
  (\itemrv_{ij}^{(1)} )} \leq \pp \frac{\delta^2}{(\frac{1}{2} -
  \delta) ( \frac{1}{2}+\delta)}.
\end{align*}
\\
 {\underline{Case III}: Suppose that $i = a,\ j = b$.} We then have
\begin{align*}
\kl{ \mprob^a(\itemrv_{ij}^{(1)}) }{ \mprob^b (\itemrv_{ij}^{(1)} )}
\leq \pp \frac{(2\delta)^2}{(\frac{1}{2} - \delta) (
  \frac{1}{2}+\delta)}.
\end{align*}
\\
Combining the bounds from all three cases, we find that the KL
divergence is upper bounded as
\begin{align*}
\frac{1}{\numrepeat} \kl{\mprob^a}{\mprob^b} \leq 2(\numitems-2)\pp
\frac{\delta^2}{(\frac{1}{2} - \delta) ( \frac{1}{2}+\delta)} + \pp
\frac{(2\delta)^2}{(\frac{1}{2} - \delta) ( \frac{1}{2}+\delta)}.
\end{align*}
Some simple algebraic manipulations yield the claimed result.

%%%%%%%%%%%%%%%%%%%%%%%%%%%%%%%%%%%%%%%%%%%%%%%%%%%%%%%%%%%%%%%%%%%%
\subsection{Proof of Corollary~\ref{cor:rank}}
\label{SecProofCorRank}

We now turn to the proof of Corollary~\ref{cor:rank}.  Beginning with
the claim of sufficiency, it is easy to see that  the
ranking is correctly recovered whenever the top $\toprank$ items are correctly recovered 
for every value of $\toprank \in [\numitems]$. Consequently, one can
apply the union bound to~\eqref{eq:topk_tail} over all values of
$\toprank \in [\numitems]$ and this gives the desired upper bound.

Now turning to the claim of necessity, we first introduce some
notation to aid in subsequent discussion. Defining the
parameter $\criticalrank \defn \min_{j \in [\numitems-1]}
(\score_{(j)} - \score_{(j+1)})$, we have shown that the lower bound
\begin{align*} 
\criticalrank \geq 8 \sqrt{\frac{\log \numitems}{\numitems \pp
    \numrepeat}}
\end{align*}
is sufficient to guarantee exact recovery of the full ranking.
Further, one must also have
\begin{align*}
\criticalrank \leq \frac{1}{\numitems - 1} \sum_{j=1}^{\numitems-1}
(\score_{(j)} - \score_{(j+1)}) = \frac{1}{\numitems - 1}
(\score_{(1)} - \score_{(\numitems)}) \leq \frac{1}{\numitems - 1}.
\end{align*}

Here we show that these two requirements are tight up to constant
factors, meaning that for any value of $\criticalrank$ satisfying
$\criticalrank \leq \frac{1}{9} \sqrt{\frac{\log \numitems}{\numitems
    \pp \numrepeat}}$ and $\criticalrank \leq
\frac{1}{9}\frac{1}{\numitems - 1}$, there are instances where
recovery of the underlying ranking fails with probability at least
$\frac{1}{70}$ for any estimator.

Consider the following ensemble of $(\numitems - 1)$ different
problems, indexed by $a \in [\numitems-1]$. For every value of $a \in
[\numitems-1]$, define a permutation $\perm^a$ of the $\numitems$
items as
\begin{align*}
\perm^a(i) =
\begin{cases}
i + 1 & \quad \mbox{if $i = a$} \\
i - 1 & \quad \mbox{if $i = a+1$} \\
i & \quad \mbox{otherwise.}
\end{cases}
\end{align*}
In words, the permutation $\perm^a$ equals the identity permutation
except for the swapping of items $a$ and $(a+1)$. Define an associated
matrix of pairwise-comparison probabilities $\Mmat^a$ as
\begin{align*}
\Mmat^a_{i j} = \half - (\perm^a(i) - \perm^a(j)) \criticalrank,
\end{align*}
and $\Mmat^a_{ji} = 1 - \Mmat^a_{ij}$. Let $\mprob^a$ denote the probabilities taken under pairwise comparisons drawn according to the model $\Mmat^a$. The condition $\criticalrank \leq \frac{1}{9} \frac{1}{\numitems - 1}$
ensures that this construction is a valid probability
distribution. One can then compute that under distribution $\mprob^a$,
the score $\score_i^a$ of any item $i$ equals
\begin{align*}
\score_i^a = \half - \big(\perm^a(i) - \frac{\numitems + 1}{2} \big)
\criticalrank.
\end{align*}
One can also verify that for any $a \in [\numitems - 1]$, and any $i
\in [\numitems-1]$, we have
\begin{align*}
\score_{\perm^a(i)}^a - \score_{\perm^a(i+1)}^a = \criticalrank,
\end{align*}
where we have used the fact that $\perm^a( \perm^a(i) ) = i$. The
requirement imposed by the hypothesis is thus satisfied.

We now use Fano's inequality~\eqref{EqnWeakFano} obtain the claimed
lower bound. In order to apply this result, we first obtain an upper
bound on the Kullback-Leibler divergence between the probability
distributions of the observed data under any pair of problems
constructed above.
\begin{lemma}
\label{lem:KLrank}
For any distinct pair $a,b \in [\numitems-1]$, we have
\begin{align*}
\kl{\mprob^a}{\mprob^b} \leq  50 \numitems \pp
  \numrepeat \criticalrank^2.
\end{align*}
\end{lemma}
\noindent See the end of this section for the proof of this claim.

Given this bound on the Kullback-Leibler divergence, the Fano
bound~\eqref{EqnWeakFano} implies that any method $\PLAINTEST$ for
identifying the true ranking has error probability
\begin{align*}
\mprob[ \TESTY \neq A] & \geq 1 - \frac{ 50 \numitems \pp \numrepeat
  \criticalrank^2 + \log 2}{\log (\numitems - 1)} \geq \frac{1}{70},
\end{align*}
where the final inequality holds whenever $\criticalrank \leq
\frac{1}{9} \sqrt{\frac{\log \numitems}{\numitems \pp \numrepeat}}$
and $\numitems \geq 9$. \\

\noindent The only remaining detail is the proof of
Lemma~\ref{lem:KLrank}.

%%%%%%%%%%%%%%%%%%%%%%%%%%%%%%%%%%%%%%%%%%%%%%%%%%%%%%%%%%%%%%%%%%%%%%%%

\subsubsection{Proof of Lemma~\ref{lem:KLrank}} 

Since the distributions $\mprob^a$ and $\mprob^b$ are formed by
components that are independent across edges $i > j$ and repetitions
$\repvar \in [\numrepeat]$, we have
\begin{align*}
\kl{ \mprob^a}{\mprob^b} = \sum_{\repvar \in [\numrepeat]} \sum_{1
  \leq i < j \leq \numitems} \kl{ \mprob^a( \itemrv_{ij}^{(\repvar)}
  )}{ \mprob^b (\itemrv_{ij}^{(\repvar)})} \; = \; \numrepeat \sum_{1
  \leq i < j \leq \numitems} \kl{ \mprob^a( \itemrv_{ij}^{(1)} )}{
  \mprob^b (\itemrv_{ij}^{(1)})},
\end{align*}
where the second equality follows since the $\numrepeat$ trials are
all independent and identically distributed.  

We now evaluate each individual term in right hand side of the above
equation. Consider any $i, j \in [\numitems]$.  We divide our analysis
into three disjoint cases:\\
{\underline{Case I}: Suppose that $i,j \in [\numitems] \backslash
  \{a,a+1,b,b+1\}$.}  The distribution of $\itemrv_{ij}^{(1)} $ is
identical across the distributions $\mprob^a$ and $\mprob^b$. As a
result, we find that
\begin{align*}
\kl{ \mprob^a(\itemrv_{ij}^{(1)}) }{ \mprob^b
  (\itemrv_{ij}^{(1)} )} = 0.
\end{align*}
\\
{\underline{Case II}: Alternatively, suppose $i \in \{a,a+1,b,b+1\}$ and $j \in [\numitems] \backslash
  \{a,a+1,b,b+1\}$ or if $j \in \{a,a+1,b,b+1\}$ and $i \in [\numitems] \backslash
  \{a,a+1,b,b+1\}$. Then we have
\begin{align*}
\kl{ \mprob^a(\itemrv_{ij}^{(1)}) }{ \mprob^b
  (\itemrv_{ij}^{(1)} )} \leq 5 \pp \criticalrank^2,
\end{align*}
where we have used the fact that $\mprob^a(\itemrv_{ij}^{(1)})$ and $\mprob^b(\itemrv_{ij}^{(1)})$ both take values in $[\frac{7}{18}, \frac{11}{18}]$ since $\criticalrank \leq \frac{1}{9} \frac{1}{\numitems-1}$.\\
{\underline{Case III}: Otherwise, suppose that both $i,j \in \{a,a+1,b,b+1\}$. Then we have
\begin{align*}
\kl{ \mprob^a(\itemrv_{ij}^{(1)}) }{ \mprob^b
  (\itemrv_{ij}^{(1)} )} \leq 20 \pp \criticalrank^2.
\end{align*}
\\
Combining the bounds from the three cases, we find that the KL
divergence is upper bounded as
\begin{align*}
\frac{1}{\numrepeat} \kl{\mprob^a}{\mprob^b} \leq 40(\numitems-4)\pp
\criticalrank^2 + 240 \pp \criticalrank^2 \leq 50 \numitems \pp
\criticalrank^2,
\end{align*}
where we have used the assumption $\numitems \geq 9$ to obtain the final inequality.

%%%%%%%%%%%%%%%%%%%%%%%%%%%%%%%%%%%%%%%%%%%%%%%%%%%%%%%%%%%%%%%%%%%%

\subsection{Proof of Theorem~\ref{thm:topkham}}
\label{SecProofThmTopkham}

We now turn to the proof of Theorem~\ref{thm:topkham}, beginning
with part (a).

%%%%%%%%%%%%%%%%%%%%%%%%%%%%%%%%%%%%%%%%%%%%%%%%%%%%%%%%%%%%%%%%%%%%%%%%%%%%

\subsubsection{Proof of part (a)}

Without loss of generality, we can assume that the true underlying
ranking is the identity ranking, that is, item $i$ is ranked at position $i$ for every $i \in [\numitems]$.  Given the
the lower bound $\KEYCON \geq 8$ is satisfied, Theorem~\ref{thm:topk}
ensures that with probability at least $1 - \numitems^{-16}$, the
counting estimator $\toptilde$ ranks every item in
$\{1,\ldots,\toprank - \hamthr\}$ higher than every item in the set
$\{\toprank + \hamthr + 1, \ldots, \numitems \}$.  Thus, we are
guaranteed that either $\toptilde \subseteq [\toprank + \hamthr]$
and/or $[\toprank - \hamthr] \subseteq \toptilde$. One can verify
either case leads to $\cardinality{\toptilde \cap [\toprank]} \geq
\toprank - \hamthr$, thereby proving the claimed result.

%%%%%%%%%%%%%%%%%%%%%%%%%%%%%%%%%%%%%%%%%%%%%%%%%%%%%%%%%%%%%%%%%%%%%%%%

\subsubsection{Proof of part (b)}

We assume without loss of generality that $\toprank \leq
\frac{\numitems}{2}$.  (Otherwise, one can equivalently study the
problem of recovering the last $\toprank$ items.)  Since the case
$\hamthr = 0$ is already covered by Theorem~\ref{thm:topk}(b), we may
also assume that $\hamthr \geq 1$.

The proof involves construction of $\packnum \geq 1$ sets of
probability matrices $\{\Mmat^a\}_{a \in [\packnum]}$ of the pairwise comparisons with the
following two properties:
\begin{enumerate}[label = (\roman*), leftmargin = *]
\item For every $a \in [\packnum]$, let $\topset_{\toprank}^a
  \subseteq [\numitems]$ denote the set of the top ${\toprank}$ items
  under the $a^{th}$ set of distributions. Then for every
  $\toprank$-sized set $\topset \in [\numitems]$,
\begin{align*}
\sum_{a = 1}^{\packnum} \indicator{
  \dham(\topset,\topset_{\toprank}^a) \leq 2 \hamthr } \leq 1.
\end{align*}
\item If the underlying distribution $a$ is chosen uniformly at random
  from this set of $\packnum$ distributions, then any estimator that
  attempts to identify the underlying distribution $a \in [\packnum]$
  errs with probability at least $\frac{1}{7}$.
\end{enumerate}
Now consider any estimator $\tophat$ for identifying the top
$\toprank$ items $\topstar$. Given property (i), whenever the
estimator is successful under the Hamming error requirement
$\dham(\tophat,\topstar)\leq 2 \hamthr$, it must be able to uniquely
identify the index $a \in [\packnum]$ of the underlying distribution
of pairwise comparison probabilities. However, property (ii) mandates
that any estimator for identifying the underlying distribution errs
with a probability at least $\frac{1}{7}$. Assuming that such sets of
probability distributions satisfying these two properties exist,
putting these results together yields the claimed result.

We now proceed to construct probability distributions satisfying the
two aforementioned properties. Consider any positive number
$\critical_0$ satisfying the upper bound
\begin{align}
\label{EqnDefnCrit0Ham}
\critical_0 \leq \frac{1}{14} \sqrt{\frac{\hamconexp \hamconmul \log
    \numitems}{\numitems \pp \numrepeat}}.
\end{align} 
The $\packnum$ matrices $\{ \Mmat^a\}_{a \in [\packnum]}$ of probability distributions we construct differ only in a permutation of their rows and columns, and modulo this permutation, have identical values. In other words, these $\packnum$ distributions differ only in the identities of the $\numitems$ items and the values of the
pairwise-comparison probabilities $\Mmat^a_{(i) (j)}$ among the
ordered sequence of the $\numitems$ items are identical across all
distributions $a \in [\packnum]$.

For any ordering $(1),\ldots,(\numitems)$ of the $\numitems$ items, for every $a \in [\packnum]$, set
\begin{align}
\label{EqnHamLowerProbs}
\Mmat^a_{(i)(j)} = 
\begin{cases}
\half + \critical_0 & \qquad \mbox{if $i \in [\toprank]$ and $j \notin
  [\toprank]$}\\
\half - \critical_0 & \qquad \mbox{if $i \notin [\toprank]$ and $j \in
  [\toprank]$}\\
\half & \qquad \mbox{otherwise.}
\end{cases}
\end{align}
Note that the upper bound~\eqref{EqnDefnCrit0Ham} on $\critical_0$,
coupled with the assumption $\pp \geq \sqrt{\frac{\log \numitems}{2
    \numitems \numrepeat}}$, ensures that $\critical_0 < \frac{1}{3}$
and hence that our definition~\eqref{EqnHamLowerProbs} leads to a
valid set of probabilities. Given this construction, the scores of the
$\numitems$ items are $\score_{(1)} = \cdots = \score_{(\toprank)} =
\score_{(\toprank+1)} + \critical_0 = \cdots = \score_{(\numitems)} +
\critical_0$. The bound~\eqref{EqnDefnCrit0Ham} ensures that the
condition $\KEYCON \leq \frac{\sqrt{\hamconexp \hamconmul}}{14}$
required by the hypothesis of the theorem is satisfied.

It remains to specify the ordering of the $\numitems$ items in each
set of probability distributions. This specification relies on the
following lemma, that in turn uses a coding-theoretic result due to
Levenshtein~\cite{levenshtein1971upper}.  It applies in the regime
$2\hamthr \leq \frac{1}{1 + \hamconmul} \min\{\numitems^{1 -
  \hamconexp}, \toprank, \numitems - \toprank \}$ for some constants
$\hamconexp \in (0,1)$ and $\hamconmul \in (0,1)$, and when
$\numitems$ is larger than a $(\hamconexp,\hamconmul)$-dependent
constant.
\begin{lemma}
\label{LemFixedWtHam}
Under the previously given conditions, there exists a subset $\{b^1,
\ldots, b^\mary \} \subseteq \{0,1 \}^{\numitems/2}$ with cardinality $\mary
\geq e^{\frac{9}{10} \hamconexp \hamconmul \hamthr \log \numitems}$,
and such that
\begin{align*}
\dham(b^j, \mathbf{0}) = 2(1+ \hamconmul)\hamthr, \quad \mbox{and}
\quad \dham(b^j, b^k) > 4 \hamthr \quad \mbox{for all $j \neq k \in
  [\mary]$.}
\end{align*}
\end{lemma}
We prove this lemma at the end of this section. Given this lemma, we
now complete the proof of the theorem. Map the $\frac{\numitems}{2}$
items $\{\frac{\numitems}{2}+1, \ldots, \numitems\}$ to the
$\frac{\numitems}{2}$ bits in each of the strings given by
Lemma~\ref{LemFixedWtHam}. For each $\ell \in [e^{\frac{9}{10}
    \hamconexp \hamconmul \hamthr \log \numitems}]$, let $B_\ell$
denote the $2 (1 + \hamconmul) \hamthr$-sized subset of $\{
\frac{\numitems}{2}+1, \ldots, \numitems \}$ corresponding to the $2
(1 + \hamconmul) \hamthr$ positions equalling $1$ in the $\ell^{th}$
string. Also define sets $A_\ell = \{1, \ldots, \toprank - 2 (1 +
\hamconmul)\hamthr\}$ and $C_\ell = [\numitems] \backslash (A_\ell
\cup B_\ell)$. We note that this construction is valid since $2\hamthr
\leq \frac{1}{1 + \hamconmul} \toprank$.

We now construct $\packnum = e^{\frac{9}{10} \hamconexp \hamconmul
  \hamthr \log \numitems}$ sets of pairwise comparison probability
distributions $\Mmat^1,\ldots,\Mmat^\packnum$ and show that these sets
satisfy the two required properties. As mentioned earlier, each matrix
of comparison-probabilities $\Mmat^\ell$ takes values as given
in~\eqref{EqnHamLowerProbs}, but differs in the underlying ordering of
the $\numitems$ items. In particular, associate the set $\ell \in
[\packnum]$ of distributions to any ordering of the $\numitems$ items
that ranks every item in $A_\ell$ higher than every item in $B_\ell$,
and every item in $B_\ell$ in turn higher than every item in
$C_\ell$. Then for any $\ell$, the set of top $\toprank$ items is
given by $A_\ell \cup B_\ell$. From the guarantees provided by
Lemma~\ref{LemFixedWtHam}, for any distinct $\ell, m \in [\packnum]$,
we have $\dham(A_\ell \cup B_\ell, A_m \cup B_m ) \geq 4 \hamthr+ 1$.
This construction consequently satisfies the first required property.

We now show that the construction also satisfies the second property:
namely, it is difficult to identify the true index.  We do so using
Fano's inequality~\eqref{EqnWeakFano}, for which we denote the
probability distribution of the observations due to any matrix
$\Mmat^\ell$, $\ell \in [\packnum]$, as $\mprob^\ell$.

We first derive an upper
bound on the Kullback-Leibler divergence between any two distributions
$\mprob^\ell$ and $\mprob^m$ of the observations. Observe that $\mprob^\ell(i \succ j) \neq \mprob^m(i \succ j)$ only if
$i \in B_\ell \cup B_m$ or $j \in B_\ell \cup B_m$. In this case, we
have $\kl{\mprob^\ell(i \succ j)}{\mprob^m(i \succ j)} \leq \frac{ 4
  \critical_0^2 }{ \frac{1}{4} - \critical_0^2}$. Since both sets
$B_\ell$ and $B_m$ have a cardinality of $2 (1 + \hamconmul) \hamthr$,
aggregating over all possible observations across all pairs, we obtain
that
\begin{align}
\label{EqnHamKLBound}
\kl{\mprob^\ell}{\mprob^m} & \leq 4 (1 + \hamconmul) \hamthr \numitems
\pp \numrepeat \frac{ 4 \critical_0^2 }{ \frac{1}{4} - \critical_0^2}.
\end{align}
In the regime $\pp \geq \frac{\log \numitems}{2 \numitems \numrepeat}$
and $\critical_0 \leq \frac{1}{14} \sqrt{\frac{\hamconexp \hamconmul
    \log \numitems}{\numitems \pp \numrepeat}}$, we have $\critical_0
\leq \frac{1}{14\sqrt{2}}$. Substituting the inequality
\mbox{$\critical_0 \leq \frac{1}{14} \sqrt{\frac{\hamconexp \log
      \numitems}{\numitems \pp \numrepeat}}$} in the numerator and
\mbox{$\frac{1}{4} - \critical^2_0 \geq \frac{1}{4} -
  \big(\frac{1}{14\sqrt{2}}\big)^2$} in the denominator of the right
hand side of the bound~\eqref{EqnHamKLBound}, we find that
\begin{align*}
\kl{\mprob^\ell}{\mprob^m} & \leq \frac{3}{4} \hamconexp \hamconmul
\hamthr \log \numitems.
\end{align*}

Now suppose that we drawn $Y$ from some distribution chosen uniformly
at random from $\{\mprob^1,\ldots, \mprob^\packnum\}$.  Applying
Fano's inequality~\eqref{EqnWeakFano} ensures that any test
$\PLAINTEST$ for estimating the index $A$ of the chosen distribution
must have error probability lower bounded as
\begin{align*}
\mprob \big[\TESTY \neq A] & \geq \left(1 - \frac{ \frac{3}{4}
  \hamconexp \hamconmul \hamthr \log \numitems + \log 2}{ \frac{9}{10}
  \, \hamconexp \hamconmul \hamthr \log \numitems} \right) \geq
\frac{1}{7}.
\end{align*}
Here the final inequality holds as long as $\numitems$ is larger than
some universal constant.

%%%%%%%%%%%%%%%%%%%%%%%%%%%%%%%%%%%%%%%%%%%%%%%%%%%%%%%%%%%%%%%%%%

\subsubsection{Proof of Lemma~\ref{LemFixedWtHam}}
\label{SecProofLemFixedWtHam}

We divide the proof into two cases depending on the value of
$\hamthr$.

\underline{Case I: $\hamthr \geq \frac{1}{2 \hamconexp \hamconmul}$:}
Let $\mary$ denote the number of binary strings of length $m_0$ such that
each has a Hamming weight $w_0$ and each pair has a Hamming distance
at least $d_0$.  It is
known~\cite{levenshtein1971upper,jiang2004asymptotic} that $\mary$ can be
lower bounded as:
\begin{align*}
\packnum & \geq \frac{ {m_0 \choose w_0} }{ \sum_{i=0}^{\lfloor
    \frac{d_0-1}{2} \rfloor} {w_0 \choose j} {m_0 - w_0 \choose j} }
\geq \frac{ \big(\frac{m_0 }{ w_0}\big)^{w_0} }{ \frac{ d_0+1}{2}
  \big( \frac{e w_0 }{ \min\{d_0, w_0\}/2} \big)^{\min\{d_0, w_0\}/2}
  \big( \frac{e m_0 }{ \min\{d_0, m_0\}/2} \big)^{\min\{d_0, m_0\}/2}
}.
\end{align*}
Note that for the setting at hand, we have $m_0 =
\frac{\numitems}{2}$, $w_0 = 2 (1 + \hamconmul)\hamthr$ and $d_0 = 4
\hamthr + 1$. Since \mbox{$\hamconexp \in (0,1)$} and
\mbox{$\hamconmul \in (0,1)$,} we have the chain of inequalities 
\begin{align*}
w_0 < d_0 \leq 4 \numitems^{1 - \hamconexp} \stackrel{(i)}{<}
\frac{\numitems}{2} = m_0, 
\end{align*}
where the inequality $(i)$ holds when $\numitems$ is large
enough. These relations allow for the simplification:
\begin{align*}
\log \packnum & \geq \log \left \{ \frac{ \big(\frac{m_0 }{
    w_0}\big)^{w_0} }{ \frac{ d_0+1}{2} \big( \frac{e w_0 }{ w_0/2}
  \big)^{w_0/2} \big( \frac{e m_0 }{ d_0/2} \big)^{d_0/2} } \right \}
\\
& = (w_0 - d_0/2) \log m_0 - w_0 \log w_0 + \frac{d_0}{2} \log d_0 -
\frac{d_0 + w_0}{2} \log (2e) - \log ((d_0+1)/2).
\end{align*}
Substituting the values of $w_0$, $d_0$ and $m_0$ and then simplifying
yields
\begin{align*}
\log \packnum & \geq (2 \hamconmul \hamthr - \half) \log
\frac{\numitems}{2} - 2 (1 + \hamconmul) \hamthr \log (2
(1+\hamconmul)\hamthr) + (2\hamthr + \half) \log (4\hamthr + 1) \\
& \qquad - (((3 + \hamconmul)\hamthr) + \half) \log (2e) - \log
(2\hamthr+1)\\
& \geq (2\hamconmul \hamthr - \half) \log \frac{\numitems}{2} -
2\hamconmul \hamthr \log (2(1 + \hamconmul) \hamthr) - \plaincon_1'
\hamthr,
\end{align*}
where $\plaincon_1'$ is a constant whose value depends only on
$(\hamconexp,\hamconmul)$. In the regime \mbox{$\frac{1}{\hamconexp
    \hamconmul} \leq 2 \hamthr \leq \frac{\numitems^{1 -
      \hamconexp}}{1+\hamconmul}$,} some algebraic manipulations then
yield
\begin{align*}
\log \packnum & \geq (2 \hamconexp \hamconmul \hamthr - \half) \log
\frac{\numitems}{2} - \plaincon_1' \hamthr \geq \hamconexp \hamconmul
\hamthr (\log \numitems - \log 2 - \plaincon_1') \geq \frac{9}{10}
\hamconexp \hamconmul \hamthr \log \numitems,
\end{align*}
where the final inequality holds when $\numitems$ is large enough. \\

\vtiny

\underline{Case II: $\hamthr < \frac{1}{2 \hamconexp \hamconmul}$}
Consider a partition of the $\frac{\numitems}{2}$ bits into
$\frac{\numitems}{4(1 + \hamconmul) \hamthr}$ sets of size $2(1 +
\hamconmul) \hamthr$ each. Define an associated set of
$\frac{\numitems}{4(1 + \hamconmul) \hamthr}$ sets of binary strings, each of length $\frac{\numitems}{2}$, with the $i^{th}$ string
having ones in the positions corresponding to the $i^{th}$ set in the
partition and zeros elsewhere. Then each of these strings have a
Hamming weight of $2(1+\hamconmul)\hamthr$, and every pair has a
Hamming distance at least $4(1 + \hamconmul)\hamthr > 4\hamthr$. The
total number of such strings equals
\begin{align*}
\exp\big(\log \frac{\numitems}{4(1 + \hamconmul) \hamthr} \big)
\stackrel{(i)}{\geq} \exp\big(\log \numitems - \log(\frac{2(1 +
  \hamconmul)}{\hamconexp \hamconmul} ) \big) \stackrel{(ii)}{\geq}
\exp \big( \frac{9}{10} \log \numitems) \stackrel{(iii)}{>} \exp \big( 1.8
\hamconexp \hamconmul \hamthr \log \numitems \big),
\end{align*}
where the inequalities $(i)$ and $(iii)$ are a result of operating in
the regime $\hamthr < \frac{1}{2 \hamconexp \hamconmul}$ and the
inequality $(ii)$ assumes that $\numitems$ is greater than a
$(\hamconexp,\hamconmul)$-dependent constant.

%%%%%%%%%%%%%%%%%%%%%%%%%%%%%%%%%%%%%%%%%%%%%%%%%%%%%%%%%%%%%%%%%%%%%%%%

\subsection{Proof of Theorem~\ref{ThmGeneralReq}} 

We now turn to the proof of Theorem~\ref{ThmGeneralReq}.

%%%%%%%%%%%%%%%%%%%%%%%%%%%%%%%%%%%%%%%%%%%%%%%%%%%%%%%%%%%%%%%%%%%%%%%

\subsubsection{Proof of part (a)}
For every $i \in [\numitems]$, let $(i)$ denote the item ranked $i$
according to their latent scores, as defined in
equation~\eqref{eq:defn_scores}. Recall from the proof of
Theorem~\ref{thm:topk} that for any $\winstart < \winend \in
[\numitems]$, the condition
\begin{align*}
\score_{(\winstart)} - \score_{(\winend)} \geq 8\sqrt{\frac{ \log
      \numitems}{\numitems \pp \numrepeat}}
\end{align*}
ensures that with probability at least $1 - \numitems^{-14}$, every
item in the set $\{(1),\ldots,(\winstart)\}$ wins more comparisons
than every item in the set
$\{(\winend),\ldots,(\numitems)\}$. Consequently, if the set
$\toptilde$ contains any item in $\{(\winend),\ldots,(\numitems)\}$,
then it must contain the entire set $\{(1),\ldots,(\winstart)\}$. In
other words, at least one of the following must be true: either
$\{(1),\ldots,(\winstart)\} \subseteq \toptilde$ or $\toptilde
\subseteq \{(1),\ldots,(\winend - 1)\}$. Consequently, in the regime
$\winend = \toprank + \alloweditem - \winstart + 1$ for any $1 \leq
\winstart \leq \toprank$ and $\winstart \leq \alloweditem \leq
\numitems$, we have that
\begin{align}
\label{EqnGeneralContainment}
\cardinality{\toptilde \cap \{(1),\ldots, (\alloweditem)\} } \geq \winstart.
\end{align}
Now consider any $\varallowedset \in [\numallowedsets]$ that satisfies the condition
\begin{align*}
\min \limits_{\varalloweditem \in [\toprank]} (\score_{(\varalloweditem)} - \score_{(\toprank + \alloweditem_\varalloweditem^\varallowedset - \varalloweditem + 1)}) \geq 8 \sqrt{\frac{\log \numitems}{\numitems \pp \numrepeat}}.
\end{align*}
For any $\varalloweditem \in [\toprank]$, setting $\winstart = \varalloweditem$ and $\winend = (\toprank + \alloweditem_\varalloweditem^\varallowedset - \varalloweditem + 1)$ in~\eqref{EqnGeneralContainment}, and applying the union bound over all values of $\varalloweditem \in [\toprank]$ yields that
\begin{align*}
\cardinality{\toptilde \cap \{(1),\ldots, (\alloweditem_\varalloweditem^\varallowedset)\} } \geq \varalloweditem \quad \mbox{for every } \varalloweditem \in [\toprank],
\end{align*}
with probability at least $1 - \numitems^{-13}$. Consequently, we have that
\begin{align*}
\mprob \big( \toptilde \in \monotone( \allowedset_\varallowedset) \big) \geq 1 - \numitems^{-13},
\end{align*}
completing the proof of the claim.

%%%%%%%%%%%%%%%%%%%%%%%%%%%%%%%%%%%%%%%%%%%%%%%%%%%%%%%%%%%%%%%%%%%%%%%%%%%

\subsubsection{Proof of part (b)}

In the regime $\alloweditem_{\genconmul \toprank}^{\varallowedset}
\leq \frac{\numitems}{2}$ for every $\varallowedset \in
     [\numallowedsets]$, it suffices to show that any estimator
     $\tophat$ will incur an error lower bounded as
\begin{align*}
\mprob \big( \cardinality{ \tophat \cap
  \{(1),\ldots,({\numitems}/{2})\} } < \genconmul \toprank \big) \geq
\frac{1}{15},
\end{align*}
where $(i)$ denotes the item ranked $i$ according to their latent
scores according to equation~\eqref{eq:defn_scores}.

Our proof relies on the result and proof of the Hamming error case
analyzed in Theorem~\ref{thm:topkham}(b). To this end, let us set the
parameter $\hamthr$ of Theorem~\ref{thm:topkham}(b) as $\hamthr = 2(1
- \genconmul)\toprank$. We claim that this value of $\hamthr$ lies in
the regime $\hamthr \leq \frac{1}{2(1+\hamconmul)} \min\{\toprank,
\numitems - \toprank , \numitems^{1 - \hamconexp} \}$ for some values
$\hamconexp \in (0,1)$ and $\hamconmul \in (0,1)$, as required by
Theorem~\ref{thm:topkham}(b). This claim follows from the fact that
\begin{align*}
\hamthr = 2(1 - \genconmul) \toprank \leq \frac{1}{2(1 + \hamconmul)}
\toprank,
\end{align*}
for $\hamconmul = \min \{ \frac{1}{4(1 - \genconmul)} - 1, \half \}
\in (0,1)$. Furthermore,
\begin{align*}
\hamthr = 2(1 - \genconmul)\toprank \stackrel{(i)}{\leq}
\frac{\numitems^{1 - \genconexp}}{4} \stackrel{(ii)}{\leq} \frac{1}{2(
  1 + \hamconmul)} \numitems^{1 - \hamconexp}
\end{align*}
for $\hamconexp = \frac{9}{10} \genconexp \in (0,1)$, where $(i)$ is a
result of our assumption $8(1 - \genconmul) \toprank \leq \numitems^{1
  - \genconexp}$ and $(ii)$ holds when $\numitems$ is large
enough. This assumption also implies that $\toprank \leq \numitems -
\toprank$ for a large enough value of $\numitems$. We have now
verified operation in the regime required by
Theorem~\ref{thm:topkham}(b).

The construction in the proof of Theorem~\ref{thm:topkham} is based on
setting 
\begin{align*}
\score_{(1)} = \cdots \score_{(\toprank)} = \score_{(\toprank
  + 1)} + \critical_0 = \cdots = \score_{(\numitems)} + \critical_0,
\end{align*}
for any real number $\critical_0$ in the interval $\Big( 0, \;
\frac{1}{14} \sqrt{ \frac{\hamconexp \hamconmul \log
    \numitems}{\numitems \pp \numrepeat}} \quad \Big]$.  This
condition is also satisfied in our construction due to the assumed
upper bound \mbox{$\KEYCON \leq \frac{1}{15} \sqrt{\genconexp
    \min\big\{\frac{1}{4(1-\genconmul)-1}, \half} \big\}$.}
Consequently, the result of Theorem~\ref{thm:topkham}(b) implies that
in this setting, any estimator $\tophat$ will incur a Hamming error
greater than $\hamthr = 2( 1 - \hamconmul) \toprank$ with probability
at least $\frac{1}{7}$, or equivalently,
\begin{align*}
\mprob \big( \cardinality{ \tophat \cap \{(1),\ldots,(\toprank)\} } <
(2\genconmul - 1) \toprank \big) \geq \frac{1}{7}.
\end{align*}
Under this event, the estimator $\tophat$ contains at most
$(2\genconmul - 1)\toprank - 1$ items from the set of top $\toprank$
items. In order to ensure it gets at least $\genconmul \toprank$ items
from $\{(1),\ldots,(\numitems/2)\}$, the remaining $2(1 - \genconmul)
\toprank + 1$ chosen items must have at least $(1 -
\genconmul)\toprank + 1$ items from
$\{(\toprank+1),\ldots,(\numitems/2)\}$. However, in the construction,
items $(\toprank+1),\ldots,(\numitems)$ are indistinguishable from
each other, and hence by symmetry these $2( 1 - \genconmul)\toprank +
1$ chosen items must contain at least $(1 - \genconmul)\toprank + 1$
items from the set $\{(\numitems/2+1),\ldots,(\numitems)\}$ with
probability at least $\half$. Putting these arguments together, we
obtain that under this construction, any estimator $\tophat$ has error
probability lower bounded as
\begin{align}
\label{EqnGenLowerAlmost}
\mprob \big( \cardinality{ \tophat \cap
  \{(1),\ldots,({\numitems}/{2})\} } < \genconmul \toprank \big) \geq
\frac{1}{14}.
\end{align}

It remains to deal with a subtle technicality. The construction above
involves items \mbox{$(\toprank+1), \ldots, (\numitems)$} with
identical scores. Recall that in the definition of the user-defined
requirement, in case of multiple items with identical scores, we
considered the choice of either of such items as valid. The following
lemma helps overcome this issue. In order to state the lemma, we
define $\matsnorm{\wt}{\infty} \defn \max_{(i,j) \in [\numitems]^2}
|\wt_{ij}|$ for a matrix $\wt \in \real^{\numitems \times \numitems}$.
\begin{lemma}
\label{LemAlmostIsFine}
Consider any two $(\numitems \times \numitems)$ matrices $\Mmat^a$ and
$\Mmat^b$ of pairwise probabilities such that
\begin{subequations}
\begin{align}
\label{EqnVBR}
\matsnorm{ \Mmat^a- \Mmat^b}{\infty} \leq \epsAlmost, \quad
\matsnorm{\Mmat^a}{\infty} \geq \epsAlmost, \mbox{ and }
\matsnorm{\Mmat^b}{\infty} \geq \epsAlmost
\end{align}
for some $\epsAlmost \in [0,1]$. Then for any $\toprank$-sized sets of
\emph{items} $\allowedset_1,\ldots,\allowedset_\numallowedsets
\subseteq [\numitems]$, and any estimator $\tophat$, we have
\begin{align}
\label{EqnVBRFinal}
\mid \mprob_{\Mmat^a}(\tophat \in
\{\allowedset_1,\ldots,\allowedset_\numallowedsets \} ) -
\mprob_{\Mmat^b}(\tophat \in
\{\allowedset_1,\ldots,\allowedset_\numallowedsets \} ) \mid \leq
6^{\numitems^2 \numrepeat} \epsAlmost.
\end{align}
\end{subequations}
\end{lemma}
\noindent
See Section~\ref{SecProofLemAlmostIsFine} for the proof of this
claim. \\

Now consider an $(\numitems \times \numitems)$ pairwise probability
matrix $\Mmat'$ whose entries takes values
\begin{align*}
\wt'_{(i)(j)} =
\begin{cases}
\half + \critical_0  + \epsAlmost & \qquad \mbox{if $i \in [\toprank]$ and $j \in [\numitems] \backslash [\numitems/2]$} \\
\half + \critical_0 & \qquad \mbox{if $i \in [\toprank]$ and $j \in
  [\numitems/2] \backslash [\toprank]$} \\
\half + \epsAlmost & \qquad \mbox{if $i \in
  [\numitems/2]\backslash[\toprank]$ and $j \in [\numitems] \backslash
  [\numitems/2]$} \\
\half & \qquad \mbox{otherwise}, 
\end{cases}
\end{align*}
and $\wt'_{ji} = 1 - \wt'_{ij}$, whenever $i \leq j$. Set $\epsAlmost = 7^{-\numitems^2 \numrepeat}$.

One can verify that under the probability matrix $\Mmat'$, the scores
of the $\numitems$ items satisfy the relations 
\begin{align*}
\score_{(1)} = \cdots = \score_{(\toprank)} = \score_{(\toprank+1)} +
\critical_0 = \cdots = \score_{(\numitems/2)} + \critical_0 =
\score_{(\numitems/2+1)} + \critical_0 + \epsAlmost = \cdots =
\score_{(\numitems)} + \critical_0 + \epsAlmost.
\end{align*}
The set of items $\{(1),\ldots,(\numitems/2)\}$ are thus explicitly
distinguished from the items
$\{(\numitems/2+1),\ldots,(\numitems)\}$. We now call upon
Lemma~\ref{LemAlmostIsFine} with $\Mmat^a = \Mmat'$, and $\Mmat^b$ as
the matrix of probabilities constructed in the proof of
Theorem~\ref{thm:topkham}, where both sets have the same ordering of
the items. This assignment is valid given that $\critical_0 <
\frac{1}{3}$ and $\epsAlmost = 7^{-\numitems^2
  \numrepeat}$. Lemma~\ref{LemAlmostIsFine} then implies that any
estimator that is $\allallowedsets$-respecting with probability at
least $1-\frac{1}{15}$ under $\Mmat^b$ must also be
$\allallowedsets$-respectiin with probability at least
$1-\frac{1}{14.5}$ under $\Mmat^a$. But by
equation~\eqref{EqnGenLowerAlmost}, the latter condition is
impossible, which implies our claimed lower bound.

%%%%%%%%%%%%%%%%%%%%%%%%%%%%%%%%%%%%%%%%%%%%%%%%%%%%%%%%%%%%%%%%%%%%%%%%%%

\subsubsection{Proof of Lemma~\ref{LemAlmostIsFine}}
\label{SecProofLemAlmostIsFine}

Let $\mprob^a$ and $\mprob^b$ denote the probabilities induced by the
matrices $\Mmat^a$ and $\Mmat^b$ respectively.  Consider any fixed
observation $\obs_1 \subseteq \{0,1,\erased\}^{\numrepeat(\numitems
  \times \numitems)}$, where $\erased$ denotes the absence of an
observation. Given the bounds~\eqref{EqnVBR}, some algebra leads to
\begin{align}
\label{EqnAlmostSandwich}
\mid \mprob^a (\obs = \obs_1) - \mprob^b(\obs = \obs_1) \mid \leq
2^{\numitems^2 \numrepeat} \epsAlmost,
\end{align}
where $\mprob^a(\obs = \obs_1)$ and $\mprob^b(\obs = \obs_1)$ denote
the probabilities of observing $\obs_1$ under $\mprob^a$ and
$\mprob^b$, respectively.

Now consider any estimator $\tophat$, which is permitted to be
randomized. Let $\packnum \leq 3^{\numitems^2 \numrepeat}$ denote the
total number of possible values of the observation $\obs$, and let $\{
\obs_1,\ldots,\obs_\packnum\} = \{0,1,\erased\}^{\numrepeat(\numitems
  \times \numitems)}$ denote the set of all possible valid values of
the observation. For each $i \in [\packnum]$, let $q_i \in [0,1]$
denote the probability that the estimator $\tophat$ succeeds in
satisfying the given requirement when the data observed equals
$\obs_i$. (Recall that the given requirement is in terms of the actual
items and not their positions.) Then we have
\begin{align*}
\big| \mprob^1 ( \tophat \in
\{\allowedset_1,\ldots,\allowedset_\numallowedsets\}) - \mprob^2 (
\tophat \in \{\allowedset_1,\ldots,\allowedset_\numallowedsets\})
\big| & = \big| \sum_{i =1}^{\packnum} \mprob^1 (\obs = \obs_i)q_i -
\sum_{i =1}^{\packnum} \mprob^2(\obs = \obs_i)q_i \big| \\
& \leq \sum_{i=1}^{\packnum} \mid \mprob^1 (\obs = \obs_i) -
\mprob^2(\obs = \obs_i) \mid q_i \\
& \stackrel{(i)}{\leq} \sum_{i=1}^{\packnum} 2^{\numitems^2
  \numrepeat} \epsAlmost q_i
\stackrel{(ii)}{\leq} 6^{\numitems^2 \numrepeat} \epsAlmost,
\end{align*}
as claimed, where step (i) follows from our earlier
bound~\eqref{EqnAlmostSandwich} and step (ii) uses the bound $\packnum
\leq 3^{\numitems^2 \numrepeat}$.

%%%%%%%%%%%%%%%%%%%%%%%%%%%%%%%%%%%%%%%%%%%%%%%%%%%%%%%%%%%%%%%%%%%%%%%%%

\section{Discussion}
\label{SecDiscussion}

In this paper, we analyzed the problem of recovering the $\toprank$
most highly ranked items based on observing noisy comparisons.  We
proved that an algorithm that simply selects the items that win the
maximum number of comparisons is, up to constant factors, an
information-theoretically optimal procedure. Our results also extend
to recovering the entire ranking of the items as a simple
corollary. In empirical evaluations, this algorithm takes several
orders of magnitude lower computation time while providing higher
accuracy as compared to prior work. The results of this paper thus
underscore the philosophy of Occam's razor that the simplest answer is
often correct.

There are number of open questions suggested by our work.  The
observation model considered here is based on a random number of
observations for all pairs of comparisons.  It would be interesting to
extend our results to cases in which only specific subsets of pairs
are observed.  Moreover, we considered a random design setting where
we do not have any control over which pairs are compared. The notion
of allowable sets introduced in this paper apply to recovery of
$\toprank$-sized subsets of the items; such a formulation and
associated results may apply to recovery of partial or total orderings
of the items. A parallel line of literature
(e.g.,~\cite{kaufmann2013information, busa2013top,
  jamieson2015sparse}) studies settings in which the pairs to be
compared can be chosen sequentially in a data-dependent manner, but to
the best of our knowledge, this line of literature considers only the
metric of exact recovery of the top $\toprank$ items. It is of
interest to investigate the Hamming and allowable set recovery
problems in such an active setting.

%%%%%%%%%%%%%%%%%%%%%%%%%%%%%%%%%%%%%%%%%%%%%
\subsection*{Acknowledgements}
This work was partially supported by NSF grant CIF-31712-23800; Air
Force Office of Scientific Research grant AFOSR-FA9550-14-1-0016; and
Office of Naval Research grant DOD ONR-N00014. In addition,  NBS was
also supported in part by a Microsoft Research PhD fellowship.

%%%%%%%%%%%%%%%%%%%%%%%%%%%%%%%%%%%%%%%%%%%%%%%%%%%%%%%%%%%%%%%%%%%%%%%%%%
%% BIBLIOGRAPHY
%\printbibliography
\bibliographystyle{alpha_initials} 
\bibliography{bibtex}

\newcommand{\etalchar}[1]{$^{#1}$}
\begin{thebibliography}{BFSC{\etalchar{+}}13}

\bibitem[AS12]{ammar2012efficient}
A. Ammar and D. Shah.
\newblock Efficient rank aggregation using partial data.
\newblock In {\em ACM SIGMETRICS Performance Evaluation Review}, 2012.

\bibitem[BFSC{\etalchar{+}}13]{busa2013top}
R. Busa-Fekete, B. Szorenyi, W. Cheng, P. Weng, and E. H{\"u}llermeier.
\newblock Top-k selection based on adaptive sampling of noisy preferences.
\newblock In {\em International Conference on Machine Learning}, 2013.

\bibitem[BLM13]{boucheron2013concentration}
S. Boucheron, G. Lugosi, and P. Massart.
\newblock {\em Concentration inequalities: A nonasymptotic theory of
  independence}.
\newblock Oxford University Press, 2013.

\bibitem[BM08]{braverman2008noisy}
M. Braverman and E. Mossel.
\newblock Noisy sorting without resampling.
\newblock In {\em Proc. ACM-SIAM symposium on Discrete algorithms}, pages
  268--276, 2008.

\bibitem[BO03]{babcock2003distributed}
B. Babcock and C. Olston.
\newblock Distributed top-k monitoring.
\newblock In {\em Proceedings of the 2003 ACM SIGMOD international conference
  on Management of data}, pages 28--39, 2003.

\bibitem[BT52]{bradley1952rank}
R. Bradley and M. Terry.
\newblock Rank analysis of incomplete block designs: I. {T}he method of paired
  comparisons.
\newblock {\em Biometrika}, pages 324--345, 1952.

\bibitem[BW97]{ballinger1997decisions}
T.~P. Ballinger and N. Wilcox.
\newblock Decisions, error and heterogeneity.
\newblock {\em The Economic Journal}, 107(443):1090--1105, 1997.

\bibitem[CCF{\etalchar{+}}01]{carmel2001static}
D. Carmel, D. Cohen, R. Fagin, E. Farchi, M. Herscovici, Y.~S. Maarek, and A.
  Soffer.
\newblock Static index pruning for information retrieval systems.
\newblock In {\em ACM SIGIR conference on Research and development in
  information retrieval}, 2001.

\bibitem[Cha14]{chatterjee2014matrix}
S. Chatterjee.
\newblock Matrix estimation by universal singular value thresholding.
\newblock {\em The Annals of Statistics}, 43(1):177--214, 2014.

\bibitem[Cop51]{copeland1951reasonable}
A.~H. Copeland.
\newblock A reasonable social welfare function.
\newblock In {\em University of Michigan Seminar on Applications of Mathematics
  to the social sciences}, 1951.

\bibitem[CS15]{chen2015spectral}
Y. Chen and C. Suh.
\newblock Spectral {M}{L}{E}: Top-$ k $ rank aggregation from pairwise
  comparisons.
\newblock In {\em International Conference on Machine Learning}, 2015.

\bibitem[CT12]{cover2012elements}
T.~M. Cover and J.~A. Thomas.
\newblock {\em Elements of information theory}.
\newblock John Wiley \& Sons, 2012.

\bibitem[dB81]{de1781memoire}
J.~C. de~Borda.
\newblock M{\'e}moire sur les {\'e}lections au scrutin.
\newblock 1781.

\bibitem[DIS15]{ding2014topic}
W. Ding, P. Ishwar, and V. Saligrama.
\newblock A topic modeling approach to ranking.
\newblock In {\em Conference on Artificial Intelligence and Statistics}, 2015.

\bibitem[DM59]{davidson1959experimental}
D. Davidson and J. Marschak.
\newblock Experimental tests of a stochastic decision theory.
\newblock {\em Measurement: Definitions and theories}, pages 233--69, 1959.

\bibitem[ER60]{erdos1960evolution}
P. Erd{\H{o}}s and A. R{\'e}nyi.
\newblock On the evolution of random graphs.
\newblock {\em Publ. Math. Inst. Hung. Acad. Sci}, 5:17--61, 1960.

\bibitem[Eri13]{eriksson2013learning}
B. Eriksson.
\newblock Learning to top-k search using pairwise comparisons.
\newblock In {\em Conference on Artificial Intelligence and Statistics}, 2013.

\bibitem[FLN03]{fagin2003optimal}
R. Fagin, A. Lotem, and M. Naor.
\newblock Optimal aggregation algorithms for middleware.
\newblock {\em Journal of computer and system sciences}, 66(4):614--656, 2003.

\bibitem[HOX14]{hajek2014minimax}
B. Hajek, S. Oh, and J. Xu.
\newblock Minimax-optimal inference from partial rankings.
\newblock In {\em Advances in Neural Information Processing Systems}, 2014.

\bibitem[Hun04]{hunter2004mm}
D. Hunter.
\newblock {MM algorithms for generalized Bradley-Terry models}.
\newblock {\em Annals of Statistics}, pages 384--406, 2004.

\bibitem[IBS08]{ilyas2008survey}
I.~F. Ilyas, G. Beskales, and M.~A. Soliman.
\newblock A survey of top-k query processing techniques in relational database
  systems.
\newblock {\em ACM Computing Surveys}, 2008.

\bibitem[JKDN15]{jamieson2015sparse}
K. Jamieson, S. Katariya, A. Deshpande, and R. Nowak.
\newblock Sparse dueling bandits.
\newblock {\em arXiv preprint arXiv:1502.00133}, 2015.

\bibitem[JS08]{jagabathula2008inferring}
S. Jagabathula and D. Shah.
\newblock Inferring rankings under constrained sensing.
\newblock In {\em Advances in Neural Information Processing Systems}, 2008.

\bibitem[JV04]{jiang2004asymptotic}
T. Jiang and A. Vardy.
\newblock Asymptotic improvement of the gilbert-varshamov bound on the size of
  binary codes.
\newblock {\em IEEE Transactions on Information Theory}, 2004.

\bibitem[KK13]{kaufmann2013information}
E. Kaufmann and S. Kalyanakrishnan.
\newblock Information complexity in bandit subset selection.
\newblock In {\em Conference on Learning Theory}, pages 228--251, 2013.

\bibitem[KMS07]{kenyon2007rank}
C. Kenyon-Mathieu and W. Schudy.
\newblock How to rank with few errors.
\newblock In {\em Symposium on Theory of computing (STOC)}, pages 95--103. ACM,
  2007.

\bibitem[KS06]{kimelfeld2006finding}
B. Kimelfeld and Y. Sagiv.
\newblock Finding and approximating top-k answers in keyword proximity search.
\newblock In {\em Symposium on Principles of database systems}, 2006.

\bibitem[Lev71]{levenshtein1971upper}
V.~I. Levenshtein.
\newblock Upper-bound estimates for fixed-weight codes.
\newblock {\em Problemy Peredachi Informatsii}, 7(4):3--12, 1971.

\bibitem[Luc59]{luce1959individual}
R.~D. Luce.
\newblock {\em Individual choice behavior: A theoretical analysis}.
\newblock New York: Wiley, 1959.

\bibitem[MAEA05]{metwally2005efficient}
A. Metwally, D. Agrawal, and A. El~Abbadi.
\newblock Efficient computation of frequent and top-k elements in data streams.
\newblock In {\em Database Theory-ICDT}. 2005.

\bibitem[MGCV11]{mitliagkas2011user}
I. Mitliagkas, A. Gopalan, C. Caramanis, and S. Vishwanath.
\newblock User rankings from comparisons: Learning permutations in high
  dimensions.
\newblock In {\em Allerton Conference on Communication, Control, and
  Computing}, 2011.

\bibitem[ML65]{mclaughlin1965stochastic}
D.~H. McLaughlin and R.~D. Luce.
\newblock Stochastic transitivity and cancellation of preferences between
  bitter-sweet solutions.
\newblock {\em Psychonomic Science}, 1965.

\bibitem[MTW05]{michel2005klee}
S. Michel, P. Triantafillou, and G. Weikum.
\newblock Klee: A framework for distributed top-k query algorithms.
\newblock In {\em International conference on Very large data bases}, 2005.

\bibitem[NOS12]{negahban2012iterative}
S. Negahban, S. Oh, and D. Shah.
\newblock Iterative ranking from pair-wise comparisons.
\newblock In {\em Advances in Neural Information Processing Systems}, 2012.

\bibitem[RA14]{rajkumar2014statistical}
A. Rajkumar and S. Agarwal.
\newblock A statistical convergence perspective of algorithms for rank
  aggregation from pairwise data.
\newblock In {\em International Conference on Machine Learning}, 2014.

\bibitem[RGLA15]{rajkumar2015ranking}
A. Rajkumar, S. Ghoshal, L.-H. Lim, and S. Agarwal.
\newblock Ranking from stochastic pairwise preferences: Recovering {Condorcet}
  winners and tournament solution sets at the top.
\newblock In {\em International Conference on Machine Learning}, 2015.

\bibitem[SBB{\etalchar{+}}16]{shah2015estimation}
N.~B. Shah, S. Balakrishnan, J. Bradley, A. Parekh, K. Ramchandran, and M.~J.
  Wainwright.
\newblock Estimation from pairwise comparisons: Sharp minimax bounds with
  topology dependence.
\newblock {\em Journal on Machine Learning Research}, 2016.

\bibitem[SBGW16]{shah2015stochastically}
N.~B. Shah, S. Balakrishnan, A. Guntuboyina, and M.~J. Wainright.
\newblock Stochastically transitive models for pairwise comparisons:
  Statistical and computational issues.
\newblock In {\em International Conference on Machine Learning (ICML)}, 2016.

\bibitem[SBW16]{shah2016feeling}
N.~B. Shah, S. Balakrishnan, and M.~J. Wainwright.
\newblock Feeling the {B}ern: Adaptive estimators for {B}ernoulli probabilities
  of pairwise comparisons.
\newblock In {\em International Symposium on Information Theory}, 2016.

\bibitem[SPX14]{soufiani2014computing}
H. Soufiani, D. Parkes, and L. Xia.
\newblock Computing parametric ranking models via rank-breaking.
\newblock In {\em International Conference on Machine Learning}, 2014.

\bibitem[Thu27]{thurstone1927law}
L.~L. Thurstone.
\newblock A law of comparative judgment.
\newblock {\em Psychological Review}, 34(4):273, 1927.

\bibitem[Tsy08]{Tsybakovbook}
A. Tsybakov.
\newblock {\em Introduction to Nonparametric Estimation}.
\newblock Springer Series in Statistics. 2008.

\bibitem[Tve72]{tversky1972elimination}
A. Tversky.
\newblock Elimination by aspects: A theory of choice.
\newblock {\em Psychological review}, 79(4):281, 1972.

\bibitem[WJJ13]{wauthier2013efficient}
F. Wauthier, M. Jordan, and N. Jojic.
\newblock Efficient ranking from pairwise comparisons.
\newblock In {\em International Conference on Machine Learning}, 2013.

\end{thebibliography}

\end{document}